%% file: main.tex
\PassOptionsToPackage{dvipsnames}{xcolor}
\documentclass[11pt]{article}

\usepackage[]{acl}

\usepackage{times}
\usepackage{latexsym}

\usepackage[T1]{fontenc}

\usepackage[utf8]{inputenc}

\usepackage{microtype}

\usepackage{inconsolata}

\usepackage{paralist}
\usepackage{soul}
\usepackage{graphicx}
\usepackage{amsmath}
\usepackage{amsthm}
\usepackage{booktabs}
\usepackage{bm}
\usepackage{algorithm}
\usepackage{algorithmic}
\usepackage[switch]{lineno}
\usepackage{multirow}
\usepackage{multicol}
\usepackage{hyperref}       
\usepackage{amsfonts}       
\usepackage{nicefrac}       
\usepackage{xcolor}         
\usepackage{cleveref}       
\crefname{section}{§}{§§}
\Crefname{section}{§}{§§}
\usepackage{array}
\usepackage{longtable}
\usepackage{enumitem}
\usepackage{amssymb}
\usepackage{xspace}
\usepackage{makecell}
\usepackage{bbding}
\usepackage{boldline}
\usepackage{pifont}
\usepackage{tabularx}
\usepackage{wrapfig}
\usepackage[export]{adjustbox}
\usepackage{fvextra}
\usepackage[most,breakable]{tcolorbox}
\usepackage{caption}
\usepackage{pgfplots}
\usepackage{arydshln}
\usepackage{subcaption}
\pgfplotsset{compat=newest}
\usepgfplotslibrary{units}
\usepackage{listings}

\def\eqref#1{equation~\ref{#1}}

\def\1{\bm{1}}

\DeclareMathAlphabet{\mathsfit}{\encodingdefault}{\sfdefault}{m}{sl}
\SetMathAlphabet{\mathsfit}{bold}{\encodingdefault}{\sfdefault}{bx}{n}

\crefname{equation}{equation}{equations}
\crefname{line}{line}{lines}
\crefname{section}{\S}{\S\S}
\Crefformat{section}{#2\S#1#3}
\crefrangeformat{section}{\S\S#3#1#4--#5#2#6}
\newcommand{\ours}[0]{DA-Code\xspace}
\title{\ours: Agent Data Science Code Generation Benchmark \\
for Large Language Models}

\author{
Yiming Huang\textsuperscript{\rm 1,2}\thanks{Equal contribution.},
Jianwen Luo\textsuperscript{\rm 1,2}\footnotemark[1],
Yan Yu\textsuperscript{\rm 1}, 
Yitong Zhang\textsuperscript{\rm 1},
Fangyu Lei\textsuperscript{\rm 1,2},\\
\textbf{
Yifan Wei\textsuperscript{\rm 1,2},
Shizhu He\textsuperscript{\rm 1,2},
Lifu Huang\textsuperscript{\rm 3},
Xiao Liu\textsuperscript{\rm 4},
Jun Zhao\textsuperscript{\rm 1,2},
Kang Liu\textsuperscript{\rm 1,2,5}\thanks{Corresponding authors.}}
\\
\textsuperscript{\rm 1}
The Key Laboratory of Cognition and Decision Intelligence for Complex Systems,\\
Institute of Automation, Chinese Academy of Sciences,\\
\textsuperscript{\rm 2}
School of Artificial Intelligence, University of Chinese Academy of Sciences,\\
\textsuperscript{\rm 3}
University of California, Davis
\textsuperscript{\rm 4}
Microsoft Research Asia\\
\textsuperscript{\rm 5}
Shanghai Artificial Intelligence Laboratory
\\
}

\begin{document}
\maketitle
\begin{abstract}
We introduce \ours, a code generation benchmark specifically designed to assess LLMs on agent-based data science tasks. This benchmark features three core elements:
First, the tasks within \ours are inherently challenging, setting them apart from traditional code generation tasks and demanding advanced coding skills in grounding and planning.
Second, examples in \ours are all based on real and diverse data, covering a wide range of complex data wrangling and analytics tasks.
Third, to solve the tasks, the models must utilize complex data science programming languages, to perform intricate data processing and derive the answers.
We set up the benchmark in a controllable and executable environment that aligns with real-world data analysis scenarios and is scalable. 
The annotators meticulously design the evaluation suite to ensure the accuracy and robustness of the evaluation. 
We develop the DA-Agent baseline. Experiments show that although the baseline performs better than other existing frameworks, using the current best LLMs achieves only 30.5\% accuracy, leaving ample room for improvement.
We release our benchmark at \url{https://da-code-bench.github.io}.
\end{abstract}

\input{sections/1-Introduction}

\input{sections/2-Task_Definition}

\input{sections/3-Benchmark_Construction}
\input{sections/4-Experiment}

\input{sections/5-Related_work}
\input{sections/6-Conclusion}

\input{sections/Limitations}

\section*{Acknowledgements}
This work was supported by the National Key R\&D Program of China (No. 2022ZD0160503) and the National Natural Science Foundation of China (No.62376270).
This work was also sponsored by CCF-BaiChuan-Ebtech Foundation Model Fund.

\bibliography{main}

\appendix
\onecolumn
\clearpage
\input{appendix/examples}
\clearpage
\input{appendix/experiments_details}

\clearpage
\input{appendix/evaluation_details}

\clearpage
\input{appendix/prompts}
\clearpage
\input{appendix/trajectory_examples}

\end{document}

%% file: sections/1-Introduction.tex
\section{Introduction}

\begin{figure}[htbp]
    \centering
    \includegraphics[width=0.49\textwidth]{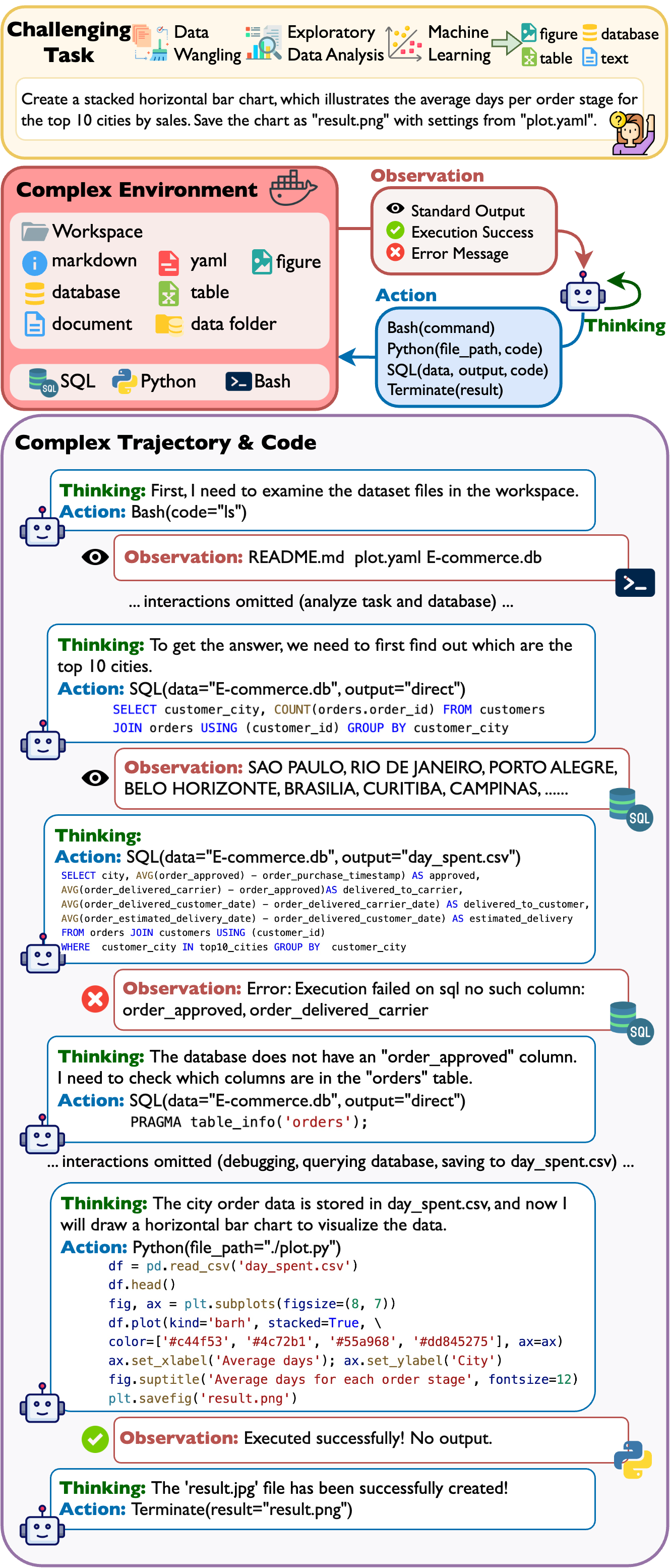}
    \caption{DA-Agent tackling an example in \ours.}
    \label{fig:main}
\end{figure}

Data science is pivotal in extracting insights from data~\citep{wang2021autods}, fundamentally shaping decision-making and knowledge discovery. 
Traditionally, this field has required high proficiency in programming and specialized knowledge, which poses significant barriers to non-experts.
However, the rapid advancement of Large Language Models~(LLMs)~\citep{openai2023gpt4, claude3, team2023gemini} has greatly enhanced their capabilities in code generation, grounding, and planning. 
This raises an intriguing question: Can LLMs become autonomous data scientists, capable of independent decision-making and solving agent data science problems?

We define the agent data science task as one that encompasses \textbf{challenging task, diverse data source, and complex solution}, as shown in Figure \ref{fig:main}. 
First, a code agent task is designed to explore data and utilize programming techniques to tackle challenging objectives, rather than simply translating explicit natural language instructions into code.
Unlike previous benchmarks like DS-1000 \citep{lai2023ds1000} and Arcade \citep{yin2023arcade}, which focus on natural language grounding \citep{xie2022unifiedskg} and transforming instructions into executable code, our approach aligns more closely with real-world programming scenarios.
Second, the data source is diverse, encompassing a variety of information and data from real programming scenarios—beyond just a notebook environment \citep{yin2023arcade} or code-completion tasks \citep{lai2023ds1000}. 
It includes different data types such as databases, spreadsheets, documents, code, and more. Moreover, these resources can be replete with distractions and extraneous information.
Third, for task solutions, data analysts must engage in a step-by-step thinking and reasoning process while programming. 
This involves inspecting data, writing minimal code, and continuous debugging. 
Additionally, they utilize Python, SQL, and Bash, which are commonly used programming languages in data science, to complete the tasks.

To achieve this goal, we introduce \ours, a benchmark for evaluating LLM data analysis ability, with carefully defined task scenarios. 
\ours contains 500 complex task examples, originating from real, challenging data analysis tasks, encompassing three main categories: \textbf{data wrangling (DW)}, \textbf{machine learning (ML)} and \textbf{exploratory data analysis (EDA)}. 
It covers the entire data analysis pipeline. 
Data wrangling includes a variety of tasks such as data loading, data cleaning, and data merging, specifically targeting raw data in files and databases.
EDA aims to gain insights and analysis using the given information and resources. It includes a wide variety of data analysis tasks using programming languages such as SQL and Python to get insights from data.
For ML tasks, how the model governs the entire ML pipeline is a promising research direction.
Each example in \ours is meticulously designed according to the standards of the code agent task mentioned above, ensuring high quality and complexity.
We meticulously design an evaluation suite for each example, and conduct red teaming experiments to validate the robustness of the evaluations.


\ours is inspired by the definitions used in intercode~\citep{yang2024intercode}. We create an interactive sandbox environment and implement \ours within this setting. This setup allows LLMs/Agents to explore the environment and engage in autonomous reasoning to complete tasks. This enables researchers to conduct their explorations in this user-friendly interactive environment.
Based on this setting, we develop a code agent baseline framework. This is capable of writing Python and SQL code and can interact with command lines, databases, and other interfaces.
As shown in Figure \ref{fig:main}, for this EDA task, the agent needs to explore all the resources provided to it, decide which files to use, write the code, and autonomously complete the task step by step.


We evaluate multiple state-of-the-art language models on \ours, and the experimental results indicate that achieving high scores on these tasks is challenging. 
For LLMs and LLM-Agents, autonomously completing real and complex data science tasks continues to be a significant challenge. 
Overall, \ours represents an agent data science benchmark with challenging real-world settings, providing valuable data resources for current LLM-agents aspiring to become data scientists.

%% file: sections/2-Task_Definition.tex
\section{Data Science Agent Task}
\label{sec:agent_task}
In this section, we introduce the data science agent task and categories of tasks for \ours.

\begin{figure*}[]
    \centering
    \includegraphics[width=0.99\textwidth]{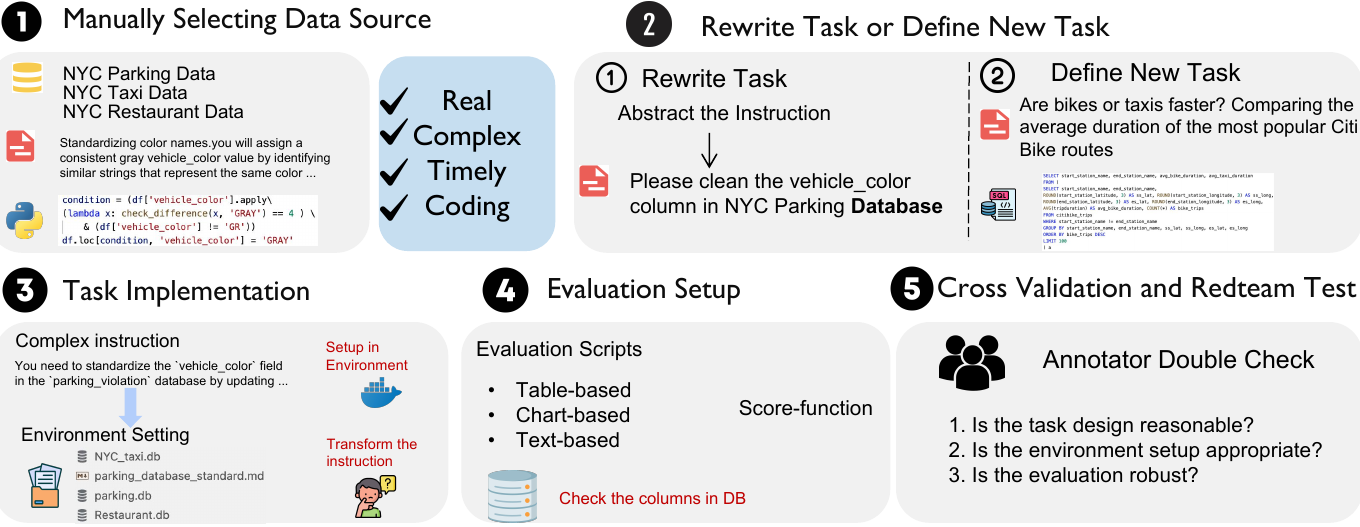}
    \caption{Annotation pipeline of \ours. See the start of Section \ref{sec:annotation_pipeline} for a detailed description.}
    \label{fig:pipeline}
\end{figure*}

\subsection{Task Definition}
\label{sec: formalization}

\setlength{\abovedisplayskip}{0pt}
\setlength{\belowdisplayskip}{0pt}


The traditional coding task can be represented as:
$$
code = f(C, I)
$$
where $code$ is the result of a function $f$ that translates contextual information 
$C$ (environmental factors, constraints) and specific instructions $I$ (requirements, tasks) into executable code.

In our coding task, interaction with the environment involves iterative code modification. The following sets are defined: \( \mathcal{S} \) (state space), \( \mathcal{A} \) (action space), \( \mathcal{O} \) (observation space), \( \mathcal{C} \) (code space), and \( \mathcal{H} \) (history space) defined as \( \mathcal{H}: \mathcal{A} \times \mathcal{C} \times \mathcal{O} \). The process can be represented as follows:
\vspace{-2pt}
\paragraph{Action Generation.} The agent takes the memory \( m_t \in \mathcal{H} \) and current state \(s_t \in \mathcal{S}\) to generate the next action \( a_{t+1} \in \mathcal{A} \) and the updated code \( code_{t+1} \in \mathcal{C} \). We use the function \( f^*: \mathcal{H} \times \mathcal{S} \rightarrow \mathcal{A} \times \mathcal{C} \) to represent this process.
$$
a_{t+1}, code_{t+1} = f^*(m_t, s_t)
$$
\paragraph{Action Execution.} The environment then interprets the agent's actions and code, executing them on the current state \( s_t \) to update the environment and obtain the new state \( s_{t+1} \). It also returns the observation \( o_{t+1} \in \mathcal{O} \). This process can be represented by the function \( g: \mathcal{A} \times \mathcal{C} \times \mathcal{S} \rightarrow \mathcal{O} \times \mathcal{S} \).
$$
o_{t+1}, s_{t+1} = g(a_{t+1}, code_{t+1}, s_t)
$$

\paragraph{Memory Update.} Subsequently, the agent's memory \( m_t \) is updated to \(m_{t+1}\) with the action \( a_{t+1} \), code \( code_{t+1} \), and observation \( o_{t+1} \). This update is constrained by the memory window size \( k \). If the updated memory length exceeds \( k \), the agent will discard the earliest memory. This process can be represented by the function \( h: \mathcal{H} \rightarrow \mathcal{H} \).
$$
m_{t+1} = h(m_t \cup \{( a_{t+1},  code_{t+1}, o_{t+1})\})
$$

This iterative process allows the agent to continually adapt and improve its code based on the evolving historical context until the agent takes an action that marks the completion of the task or reaches the maximum time set by the environment.

\input{tables/benchmark_comparisons}
\input{tables/statistic}
\subsection{Task Categories}
\ours focuses on data science tasks and categorizes them into three major categories: data wrangling, machine learning, and exploratory data analysis.
\vspace{-5pt}
\paragraph{Data Wrangling.}
Data wrangling is the process of transforming and mapping raw data from one form into another to prepare it for analysis. It involves cleaning, loading, and transforming raw data into a more usable format. This can include handling missing values, correcting errors, and merging datasets from different sources. The goal of data wrangling is to ensure that the data is consistent and easily accessible for tasks such as analytics, reporting, or machine learning applications. The examples are shown in Figure \ref{fig:pipeline} and Figure \ref{fig:data_example}.
\vspace{-5pt}
\paragraph{Exploratory Data Analysis.}
Exploratory Data Analysis is a technique used in data analysis to understand the main characteristics and get insights from a dataset. 
There are many types of EDA tasks, which require LLMs to use a combination of various Python libraries and SQL to complete complex data analysis tasks. The types of tasks included in \ours are statistical analysis, data manipulation, data insights, and data visualization. The example in Figure \ref{fig:main} is typical of this type of task.
\begin{itemize}
    \item \textbf{\textit{Data Manipulation}} is designed to perform intricate data operations using SQL and Pandas, efficiently processing large datasets. It primarily focuses on counting, summarizing data, and refining presentation formats.
    \item \textbf{\textit{Data Insights}} focuses on tasks involving real data and issues encountered in actual scenarios. These problems do not provide clear solutions, necessitating thoughtful consideration and autonomous decision-making in coding for data analysis. The results are presented primarily in tables, or alternatively, in text format, to address and answer these questions.
    \item \textbf{\textit{Visualization}} is similar to the previous category, with the unique aspect being that the results are presented in the form of charts.
    \item \textbf{\textit{Statistical Analysis}} typically requires advanced knowledge of statistics and mathematics, involving the use of mathematical indicators for analysis.
\end{itemize}


\paragraph{Machine Learning.}
ML is a crucial data analysis technique, indispensable for automating and enhancing decision-making processes.
In \ours, we select three foundational task categories: classification, regression, and clustering, from two sources: regular dataset tasks and competition tasks.
The reference solutions for these tasks require the use of corresponding ML algorithms (excluding deep learning) to complete the tasks.

The details and examples of all task types are provided in Appendix \ref{app:examples}.

%% file: tables/benchmark_comparisons.tex
\begin{table*}[t]
\centering
\setlength\tabcolsep{2pt}
\resizebox{\textwidth}{!}{
\begin{tabular}{lccccccc}
\hline
\textbf{Benchmark} & \textbf{\thead{Control.\\Exec. Env?}} & \textbf{\thead{Need \\ Planning?}} & \textbf{\thead{\#Files \\ per/ Task}}  & \textbf{Fields}                                                                      & \textbf{\# instance} & \textbf{\thead{\# Average lines \\of solutions}} & \textbf{\thead{Evaluation\\ Method}}  \\\hline
DS-1000 \citep{lai2023ds1000}            & \XSolidBrush & \XSolidBrush & 1     & Code Completion                                                                      & 1000                 & 3.6                                      & Pass@K                              \\
Arcade \citep{yin2023arcade}            & \XSolidBrush & \XSolidBrush & 1 & Notebook Completion                                                                  & 1078                 & 2.3                                      & Output Match                              \\
MLAgentBench \citep{huang2023mlagentbench}       & \checkmark & \checkmark & 4.8  & ML Project                                                                           & 13                   & -                                    & Evaluation Scripts                       \\
DA-Bench \citep{hu2024dabench}           & \checkmark   & \XSolidBrush  & 1 & CSV Analysis                                                                         & 257                  & $\sim$20                                     & Output Match                               \\ \hline
DA-Code             & \checkmark   & \checkmark & 5.7  & \begin{tabular}[c]{@{}l@{}}Data Analysis Project\end{tabular} & \textbf{500}                  & \textbf{85}                                    & Exec-based Eval          \\
    \hline

    \hline
\end{tabular}%
    }
\caption{Comparisons of several data science code generation benchmarks.}
\label{tab:benchmark_comparisons}%
\end{table*}%

%% file: tables/statistic.tex
\begin{table}[t]
  \centering
  \small
\begin{tabular}{lc}
    \toprule
    \textbf{Statistic} & \textbf{\# Task} \\
    \midrule
    \textbf{Total tasks} & \textbf{500 (100\%)} \\
    - Data Wrangling & 100 (20.0\%) \\
    - Machine Learning  & 100 (20.0\%) \\
    - Data Manipulation  & 73 (14.6\%) \\
    - Data Insights & 79 (15.8\%) \\
    - Visualization & 70 (14.0\%) \\
    - Statistical Analysis  & 78 (15.6\%) \\
    
    \midrule
    - Easy Level & 105 (22.8\%) \\
    - Medium Level & 292 (57.3\%) \\
    - Hard Level& 103 (19.9\%) \\
    \bottomrule
\end{tabular}
\caption{Data Statistics of Examples in \ours.}
\vspace{-15pt}
  \label{tab:statistic}%
\end{table}%


%% file: sections/3-Benchmark_Construction.tex
\section{\ours }
In this section, we describe the statistical information and construction pipeline of \ours.

\subsection{Challenging Tasks and Diverse Data}
In Table \ref{tab:benchmark_comparisons} and \ref{tab:statistic}, we conduct a statistical analysis where the distribution of tasks among DW, ML, and EDA is in a 1:1:3 ratio.
In contrast, \ours integrates a diverse array of agent tasks across the entire data science pipeline, covering a broad spectrum of task types and data types (Figure \ref{fig:statistic}), and result formats (such as Tables, Databases, Charts, Text, etc.). 
During the annotation process, our experienced annotators also categorize the difficulty of each task into three levels: easy, medium, and hard.
Additionally, each example in \ours involves multiple files, averaging 5.7 files per task. 
This setup more closely mirrors real data analysis scenarios.

\subsection{Complex Solution}
As shown in Table \ref{tab:benchmark_comparisons}, we curate solution codes for each example, requiring an average of 85 lines of code to complete the tasks.
Unlike previous benchmarks, \ours uses a controllable executable environment to construct complex coding tasks that require interaction with the environment, planning and coding to complete tasks. Many tasks require the use of languages like SQL and Python, which aligns closely with real-world data science analysis scenarios.  


\subsection{Evaluation Suite}
\label{sec:evaluation_setup}


We meticulously develop an accompanying evaluation suite that ensures a comprehensive and systematic assessment of the LLM-Agent performance on \ours.
\vspace{-5pt}
\paragraph{Data Standardization.} 
For each data type, we implement carefully designed scripts to extract standardized information essential for evaluation. For tables, we do not compare the entire table but instead extract specific columns. For charts, we identify plotting scripts (e.g., plot.py) and use scripts to extract both numerical data and plotting parameters, which are then stored in numpy and JSON formats. For text-based outputs, we parse them into JSON format for comparison.
\vspace{-5pt}
\paragraph{Evaluation Configure.}
The evaluation setup for each task is customized through a specific configuration, providing flexibility and ease in managing multiple tasks within the evaluation suite. Each task is uniquely identified, and necessary evaluation details, including output files, metrics, and options, are defined to meet the diverse requirements of different tasks. This structured approach enhances the efficiency and accuracy of the evaluation process.
\vspace{-5pt}
\subsection{Score Calculation}
Building on the evaluation suite, we develop a scoring methodology to assess LLM-Agent performance across various outputs, including tables, charts, and machine learning predictions. Each output type has tailored metrics for comprehensive evaluation. Detailed scoring processes are provided in Appendix \ref{app:evaluation_details}.
\vspace{-5pt}
\paragraph{Table match score.} The evaluation of tables involves comparing CSV files or databases, as well as the JSON format for text-based outputs, by matching two tables. The task instructions clearly specify the expected format, and the evaluation checks for an exact match between the task-specified columns in the predicted table \(M'\)  and the reference table  \(M\) . The score is defined as:
\[\text{Score} = 
\begin{cases} 
1, & \text{if } M' = M \\
0, & \text{otherwise}
\end{cases}\]
A perfect match results in a score of 1, while any discrepancy results in a score of 0.

\paragraph{Chart match score.} For chart evaluations, our script extracts key metadata from the predicted chart, including both the numerical data \(D'\) and plot configurations \(I'\) as specified in the task instructions. These components are then compared with the reference chart’s true values for numerical data \(D\) and plot configurations \(I\). The chart match score is calculated using the following rule:
\[\text{Score} = \begin{cases}
    1 &\; \text{if}\; D' = D\; \text{and} \; I' = I\\
    0 &\; \text{otherwise}
\end{cases}\]
\paragraph{ML normalized score.} For machine learning tasks, we utilize several standard metrics, including F1 Score, MAE, Silhouette Score, etc. Given the diverse nature of these metrics across different tasks, we apply a normalization process to map the original scores onto a 0-1 scale, ensuring consistency and comparability. The normalized score is computed using the following formula:
\[
\text{Score} = \min \left(1, \max\left(0, \frac{\hat{s} - S_{\text{baseline}}}{S_{\text{best}} - S_{\text{baseline}}}\right)\right)
\]

Here, \( \hat{s} \) represents the original score, \( S_{\text{best}} \) is the upper performance limit, and \( S_{\text{baseline}} \) is the lower bound.

\subsection{Annotation Pipeline}
\label{sec:annotation_pipeline}

We recruit ten annotators who are highly proficient in data analysis, SQL, and Python to carry out data collection and annotation. As shown in Figure \ref{fig:pipeline}, the data annotation pipeline consists of the following steps:
    \paragraph{Manually Selecting Data Source.} The data must come from actual data analysis and engineering projects. We require the datasets to be genuinely large and real, not simulated tables or texts. The data source must meet four principles: (1) real-world relevance, (2) complexity, (3) timeliness, and (4) coding intensity. We collect the most recent data sources from Kaggle, Github, and Other Web Sources. When collecting data, strive to select datasets that come with corresponding code and carefully verify their quality.
    \vspace{-5pt}
    \paragraph{Rewrite Task or Define New Task.} According to the definition of data science agent task in Section \ref{sec:agent_task}, we have two ways to define tasks. The first method, rewrite task, involves completely redefining discovered task resources by transforming explicit code instructions into abstract agent task descriptions. The second method requires manually annotating new tasks based on these discovered resources. The majority of tasks are derived from conversions of the first type.
    \vspace{-5pt}
    \paragraph{Task Implementation.} \ours tasks are set up in a specialized Sandbox environment. The most critical step involves collecting the necessary resources for data analysis. The data we gather is noisy (a task may involve multiple files), and we ensure that this noisy data serves as the initial resource. For the examples shown in Figure \ref{fig:pipeline}, which originally only involved a few databases, we create a realistic setting by providing files required in real tasks, such as ``data\_standard.md'' or ``schema.yml''. The agent needs to extract useful information from these files to complete the task. Unlike traditional language grounding tasks, this process is challenging, requiring decision-making and reasoning.
    \paragraph{Evaluation Setup.}Each example's evaluation configuration is meticulously designed based on our evaluation suite, as detailed in Section \ref{sec:evaluation_setup}. 
    \vspace{-5pt}
    \paragraph{Cross Validation and Red Team Test.} We ask annotators to perform cross-validation to ensure each example has a reasonable task design, appropriate environment setup, and robust evaluation. Additionally, they are required to conduct red team testing to determine if there are any false positives or false negatives.

%% file: sections/4-Experiment.tex
\section{DA-Agent}

\label{sec:agent}
To effectively address the challenges of the DA-Code benchmark, we develop an LLM-based agent, depicted in Figure \ref{fig:main}, which operates within a versatile and robust framework designed for dynamic interaction and execution with the environment.

\subsection{Environment}
Inspired by the work of \citet{yang2024intercode}, the environment of DA-Agent is built on the Docker platform, ensuring a consistent and isolated setting crucial for replicable and independent data science experiments. This Linux environment comes equipped with essential data science tools, including Python, SQL, Conda, and database engines.


\subsection{Action Space}

Previous approaches typically define actions in terms of editing or executing files. 
However, in our system, we innovatively combine these stages into single, streamlined actions that edit and execute code simultaneously. 
This approach not only reduces the complexity of interactions but also minimizes the number of steps required, thereby saving computational resources and enhancing model comprehension. 
Our action space is designed to efficiently manage diverse tasks, encompassing the following actions:

\setlist[itemize]{topsep=0pt, partopsep=0pt, parsep=0pt, itemsep=0pt}
\begin{itemize}
    \item \textbf{Bash(command)}: Executes single-line bash commands directly. This enables quick file and directory manipulation and system command execution, providing direct interaction with the operating system.
    
    \item \textbf{Python(save\_path, code)}: Requires path and code content of the Python code, allowing the agent to handle complex data processing tasks and utilize Python’s extensive libraries.
    
    \item \textbf{SQL(file\_path, command, output)}: Executes SQL queries by specifying the database file, SQL command, and the output format. Results can be saved to a specified file or displayed directly.
    
    \item \textbf{Terminate(output)}: Concludes the task, specifying the result file or output text. This final action ensures that results are summarized and appropriately directed, marking a clear end to the session.
\end{itemize}

This diverse range of actions equips the agent with the capabilities to handle complex tasks across different environments, making it a versatile tool in data manipulation and system operations.

\subsection{Response Mechanism}

Responses of the agent are categorized into these types based on the system's feedback to executed actions:

\setlist[itemize]{topsep=0pt, partopsep=0pt, parsep=0pt, itemsep=0pt}
\begin{itemize}

\item \textbf{Standard Output.} The output from successfully executed commands, provides direct feedback or results from the executed actions.

\item \textbf{Error Message.} In cases where execution fails, error messages are generated to aid in debugging and corrective measures.

\item \textbf{Execution Success without Output.} Some commands execute successfully without producing visible output, in which case the system simply acknowledges their successful execution.

\item \textbf{Unacceptable Action.} 
When the output format does not match the Action format, or the action is the same as the last one, please provide a different action.

\item \textbf{Execution Timeout.} 
The action execution time has exceeded the time limit.

\end{itemize}

\subsection{Memory Windows} 
To manage the context for the agent's operations, a memory window records the history of actions taken, constrained by a \textit{max history length} parameter. This parameter limits the number of previous steps the agent can recall. If the required context exceeds this limit, the history is automatically truncated to maintain efficient memory management and focus on the most recent relevant actions.

\section{Experiment and Analysis}

In this section, we present the experimental results and analysis of several LLMs evaluated using our DA-Agent baseline on \ours benchmark.

\subsection{Experiment Settings}
We experiment with state-of-the-art LLMs from open-source representatives such as 
Mixtral-8x22B \citep{jiang2024mixtral}, DeepseekCoder-V2.5 \citep{zhu2024deepseek}, Qwen2.5-72B-Instruct \citep{qwen2.5}
and closed-source ones including Claude-3-Opus~\citep{claude3} and GPT~\citep{openai2023gpt4} families.

We also compare our DA-Agent with three widely-used agent frameworks, namely OpenDevin \citep{opendevin2024}, AutoGen \citep{wu2023autogen} and X-Agent \citep{xagent2023}.

For all experiments, we employ a greedy sampling strategy with a maximum step length of 20 and a \textit{max history length} of up to 15 steps. The action execution time limitation is 300 seconds.

\input{tables/main_results}

\begin{figure}[t]
    \centering
    \includegraphics[width=0.4\textwidth]{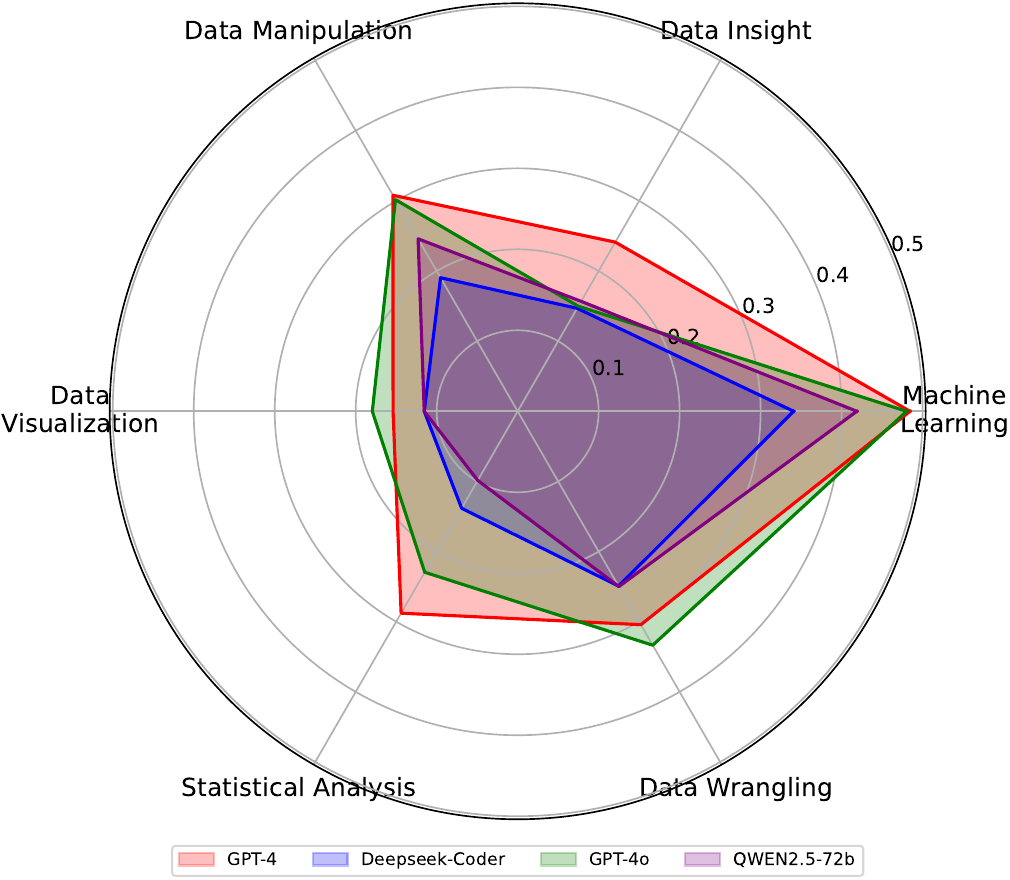}
    \caption{Detailed performance analysis of DA-Agent across various categories on DA-Code.}
    \label{fig:radio}
\vspace{-5pt}
\end{figure}

\subsection{Main Results}
\paragraph{DA-Agent with Different LLMs.}
In Table \ref{tab:main}, we compare the performances of DA-Agent based on advanced LLMs. In Figure \ref{fig:radio}, we conduct fine-grained performance statistics for the DA-Code categories. 
From the score results, we can conclude that 
1) Existing data agents are far from satisfactory in completing these data science coding tasks. The most advanced model, GPT-4, achieves only a 30.5\% score. 
2) Although closed-source models generate high-performance executable code, they have a significant gap compared to open-source LLMs in terms of overall score. 
3) We classify task difficulty into three levels. Experimental results indicate that model performance decreases with increasing difficulty, validating our grading approach. 
4) The models perform poorly on data wrangling tasks but fare better in machine learning challenges. This disparity could be linked to the training corpus, as DW and EDA tasks are less common and more complex to understand.



\paragraph{Different Agent Framework with GPT-4.} 
We randomly sample 100 tasks from \ours to create a subset DA-Code-100 for comparison with several popular code agents. As shown in Table \ref{tab:agent_framework}, our baseline DA-Agent outperforms these established agents on \ours, achieving higher score and completion rate. However, the overall performance remains modest, indicating substantial potential for enhancing coding agent frameworks to better handle complex data science tasks.

\input{tables/agent_framework}

\begin{figure}[h]
    \centering
    \includegraphics[width=0.45\textwidth]{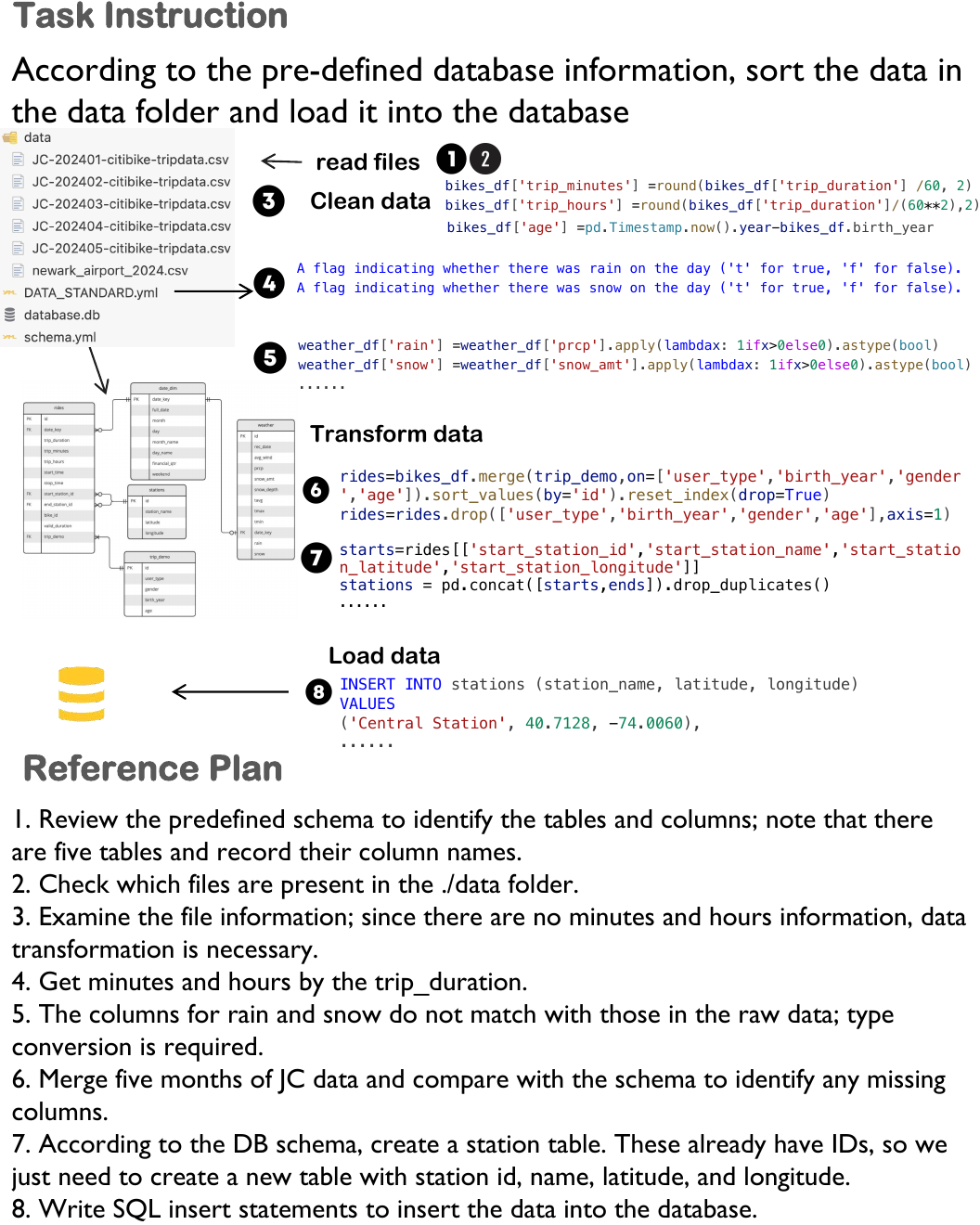}
    \caption{The task instruction and the corresponding reference plan of the \ours example.}
    \label{fig:data_example}
\end{figure}

\subsection{Ablation Study of DA-Agent}

\paragraph{Reference Plan.}  
\ours aims to assess the combined abilities of planning and grounding in LLM-agents. To further investigate the factors affecting model performance, we asked annotators to annotate the reference plan of \ours-100, as shown in Figure \ref{fig:data_example}. 
This type of instruction describes a step-by-step plan for solving a task, serving as a reference for the LLM Agents. 
Table \ref{tab:agent_framework} shows that LLMs improve with a reference plan, highlighting planning ability as a key performance factor.  
Additionally, annotating reference plans provides valuable resources for research exploration.

\paragraph{Max History Length.} 
We investigate the impact of \textit{max history length} on the performance of DA-Agent. As shown in Table \ref{tab:agent_framework}, using the \ours-100 dataset, the model's performance shows minimal change with variations.



\begin{figure}[t]
    \centering
    \includegraphics[width=0.5\textwidth]{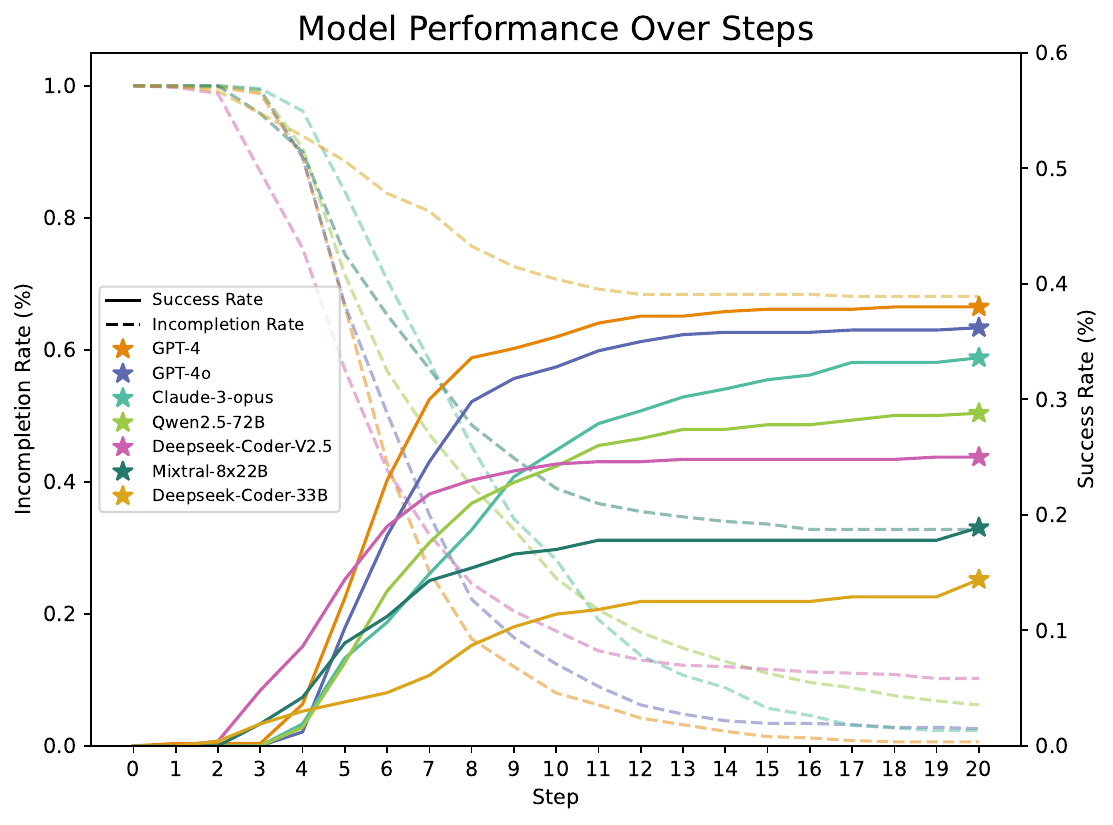}
    \caption{The success and incompletion rates of various models over steps. The incompletion rates represent the proportion of tasks not completed and the success rates indicate the proportion of tasks successfully completed within the current step. A task is considered successful if it scores above zero.}
    \label{fig:model_steps}
\end{figure}

\subsection{Step into Trajectories}
\paragraph{Task Completion Efficiency.}
We examine the success and incompletion rates across various models over a sequence of steps, as depicted in Figure \ref{fig:model_steps}. The incompletion rates consistently decrease, particularly between 5 and 15 steps, which indicates the complexity of the tasks requiring multiple steps for resolution. Success rates rise sharply within the initial 5 to 10 steps before reaching a plateau, even as incompletion rates continue to decline. This highlights a fundamental challenge: if an agent fails to grasp the task's requirements and come up with effective solution in the early steps, additional steps do not necessarily lead to successful outcomes, underscoring the limited capability of existing agents in handling complex prolonged task sequences.

\paragraph{EEEA Pattern.}

\begin{figure}[th]
    \centering
    \includegraphics[width=0.48\textwidth]{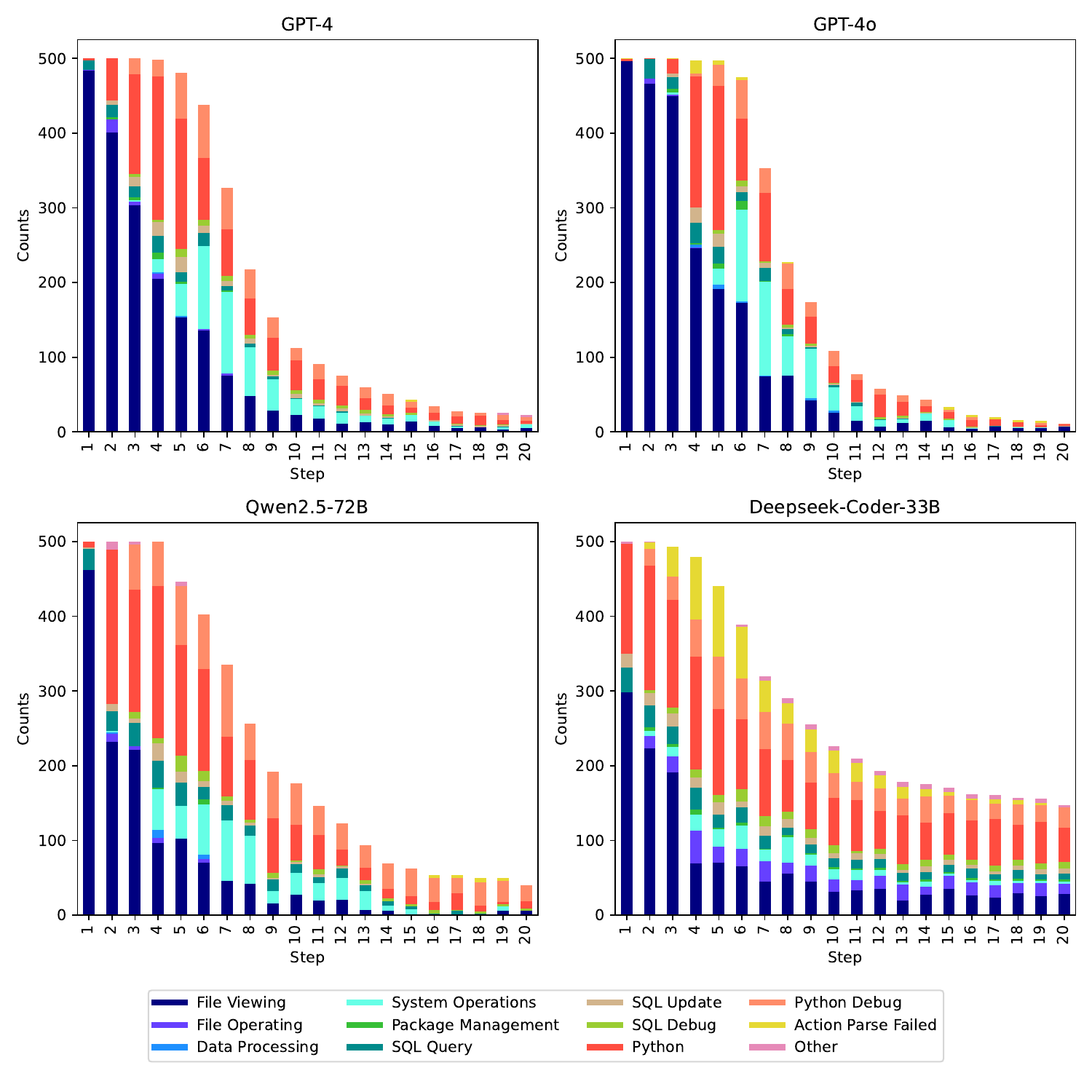}
    \caption{Action type counts of DA-Agent with different LLMs across steps.}
    \label{fig:gpt4_actions}
\end{figure}

Based on our in-depth analysis of DA-Agent's task-solving steps using different LLMs and the classification of action types detailed in Table \ref{tab:agent_actions}. As shown in Figure \ref{fig:gpt4_actions}, we observe a prevalent Exploration-Execution-Evaluation-Adjustment pattern in the agents' trajectories, which aligns well with our task scenarios. 
At the beginning of the task, barring instances of action extraction failure, each model tends to prioritize the ''File Viewing'' action to explore file contents and gain an understanding of the environment. As the task progresses, actions related to coding, such as invoking Python or executing SQL queries, become more prevalent. In the later stages of the task, higher-performing models like GPT-4 and GPT-4o gradually decrease their file operation-related actions while increasing actions associated with debugging, suggesting they may have identified potential solutions to the problems at hand.
Conversely, models with more varied performance characteristics, like Deepseek-Coder, continue focusing on file operations. This suggests these models may have a less developed ability to comprehend and adapt to the task environment effectively.

\paragraph{Error Analysis.}
From our detailed examination of the DA-Agent's actions across various models, we identify several recurring issues that contribute to errors:
\setlist[itemize]{topsep=0pt, partopsep=0pt, parsep=0pt, itemsep=0pt}
\begin{itemize}

\item \textbf{Hallucination Issues:} Agents often make incorrect assumptions about the environment, such as presuming file names and directly executing Python or SQL commands without initial exploration of available files.

\item \textbf{Inability to Follow Instructions:} This leads to many non-standard reply formats and unrecognizable actions. Some models often start with non-standard actions but adjust in later steps.

\item \textbf{Persistent Code Errors:} These errors cause failure to debug and correct issues, leading the model to become stuck in a debugging loop.

\item \textbf{Misinterpretation of Task Context:} Agents sometimes misinterpret task details, leading to premature termination and incomplete results.
\end{itemize}
\textbf{} 

In conclusion, our findings underscore the current limitations of agents in tackling complex data challenges. This study provides crucial insights that can direct the enhancement of code agent capabilities through the development of more robust and context-aware strategies.

%% file: tables/main_results.tex
\begin{table*}[htbp]
  \centering
  \renewcommand{\arraystretch}{1.05}
  \resizebox{\textwidth}{!}{
\begin{tabular}{l|ccccccc|ccc}
\hline
\textbf{\multirow{2}{*}{Model}}
& \multicolumn{7}{c|}{\textbf{Score}}                                                                                                                                 & \multicolumn{1}{c}{\multirow{2}{*}{\thead{\textbf{Completion} \\  \textbf{Rate (\%)}}}
} & \multicolumn{1}{c}{\multirow{2}{*}{\thead{\textbf{\# Avg} \\ \textbf{Steps}}}
}  & \multicolumn{1}{c}{\multirow{2}{*}{\thead{\textbf{Executable} \\ \textbf{Code (\%)}}}
} \\ \cline{2-8} & \multicolumn{1}{c}{\textbf{DW}} & \multicolumn{1}{c}{\textbf{ML}} & \multicolumn{1}{c}{\textbf{EDA}} & \textbf{Easy}& \textbf{Medium}& \textbf{Hard}& \multicolumn{1}{c|}{\textbf{Total}} & \multicolumn{1}{c}{}      & \multicolumn{1}{c}{}  & \multicolumn{1}{c}{}  \\ \hline \hline
GPT-4             & 30.4     & \textbf{48.4}      & \multicolumn{1}{l|}{\textbf{24.6}}     & 45.4 & \textbf{27.8} & \multicolumn{1}{l|}{\textbf{23.4}} & \textbf{30.5} & \textbf{99.4} & 7.3  & 76.8 \\
GPT-4o           & \textbf{33.3}  & 48.0 & \multicolumn{1}{l|}{21.3} & \textbf{46.2} & 25.6  & \multicolumn{1}{l|}{21.7} & 29.1 & 97.4 & 6.8 & \textbf{77.7} \\
Claude-3-Opus      & 29.3 & 46.8 & \multicolumn{1}{l|}{20.7} & 44.7 & 23.8  & \multicolumn{1}{l|}{19.0} & 27.6 & 97.7 & 8.9 & 75.7 \\
\hdashline
Qwen2.5-72B        & 24.9  & 41.8 & \multicolumn{1}{l|}{15.4} &  31.9 &  19.4  & \multicolumn{1}{l|}{22.3}  &  22.6    &  93.8 & 8.6 & 72.2\\
Deepseek-Coder-V2.5 & 25.1 & 34.1 & \multicolumn{1}{l|}{14.7} & 32.8  & 18.7  & \multicolumn{1}{l|}{14.1}   & 20.7 & 89.8 & 7.1 & 59.0 \\
Mixtral-8x22B      & 14.8 & 31.6 & \multicolumn{1}{l|}{10.2} & 17.6 & 16.8  & \multicolumn{1}{l|}{8.6} & 15.4 & 67.2 & 11.1 & 55.1 \\
Deepseek-Coder-33B & 9.1 & 22.1 & \multicolumn{1}{l|}{7.6} & 12.4  & 11.3  & \multicolumn{1}{l|}{7.9}   & 10.8 & 31.9 & 11.6 & 49.7 \\
\hline
\end{tabular}
    }
\caption{Experiments results of some LLMs using DA-Agent baseline. The Completion Rate (\%) represents the proportion of tasks for which the model produces results within 20 steps. \#Avg Steps indicates the number of steps the agent requires to complete these tasks. Executable Code (\%) reflects the proportion of code generated by LLMs that successfully executes.}
  \label{tab:main}%
\end{table*}%

%% file: tables/agent_framework.tex
\begin{table}[htbp]
  \centering
  \vspace{-5pt}
  \resizebox{0.48\textwidth}{!}{
\begin{tabular}{l|cc}
    \hline
    \textbf{Experiment Setting} & \textbf{\thead{Overall\\ Score}} & \textbf{\thead{Completion\\ Rate (\%)}} \\ \hline
    X-Agent \citep{xagent2023}         &   6.7         &       38.7      \\
    AutoGen \citep{wu2023autogen}       &   18.6            &     78.8          \\
    OpenDevin \citep{opendevin2024}       &   26.2          &    96.0           \\ \hdashline
    \ours          &    31.5           &     99.5          \\
    \quad - w/ Ref. Plan & 39.7 & 97.7 \\  \hdashline
    \quad - w/ \textit{max history length}=10     &    32.3          &     98.2          \\
    \quad - w/ \textit{max history length}=5  & 30.8 & 95.4  \\
    \hline
\end{tabular}
  }
  \caption{Performance comparison of different frameworks using GPT-4 on DA-Code-100.}
  \vspace{-5pt} 
  \label{tab:agent_framework}
\end{table}

%% file: sections/5-Related_work.tex
\section{Related Work}
\label{sec:related}
\vspace{-5pt}
\paragraph{Code Generation Benchmark}
As models become increasingly capable, researchers start to build increasingly difficult and general code generation benchmarks. Most coding benchmarks (e.g. SQL-Spider~\citep{yu2018spider}; Bash - NL2Bash~\citep{lin2018nl2bash}; Python - HumanEval~\citep{humaneval}; Execution-S3Eval \citep{lei2023s3eval}; Competition code generation \citep{huang2024competition}) frame the coding problem as a sequence-to-sequence problem (from instruction to code). DS-1000~\citep{lai2023ds1000} and Arcade~\citep{yin2023arcade} are pioneering works that collected high-quality examples from communities and proposed corresponding data science to define code generation tasks. 
Intercode~\citep{yang2024intercode} was the first to propose defining code generation tasks in an interactive environment. SWE-Bench~\citep{jimenez2023swebench} proposed numerous repository-level tasks, while MLAgentBench~\cite{huang2023mlagentbench} defined auto machine learning tasks in an interactive environment. 
Some researchers have also proposed benchmarks~\citep{xie2024osworld,cao2024spider2v} to explore the model's multimodal capabilities in data science and engineering. ML-Bench \cite{liu2023mlbench} focuses on machine learning bash scripts generation. DA-Bench \cite{hu2024dabench} also evaluate agents on data analysis tasks, however its task setting is not fully agentic and advanced.
Our work focuses on data science, involving real and challenging tasks that cover the full pipeline.

\paragraph{Code Agent Method}
The value of generative code models and interactive problem solving has motivated a recent proliferation of work to augment reasoning capabilities' of existing language models~\citep{yao2022react, shinn2024reflexion, chen2023selfdebug, zhang2022planning, wang2023planandsolve}. Many agent methods have been proposed to solve code generation tasks. Direct interaction of agents with the Linux command line can cause many issues~\citep{yang2024sweagent}. Several works~\citep{yang2024intercode, huang2023mlagentbench, zhang2024codeagent, yang2024sweagent} have designed special actions to standardize agent operations. 
OpenDevin~\citep{opendevin2024} is a new agent for solving coding tasks. It is based on CodeAct~\citep{wang2024codeact}, a framework that consolidates LLM agents’ actions into a unified code action space. 
We introduce DA-Agent, a competitive agent framework designed for solving coding tasks, based on the DA-Code environment.


%% file: sections/6-Conclusion.tex
\section{Conclusion}
\vspace{-5pt}
\label{sec:conclusion}

We introduce \ours, a challenging benchmark designed for agent-based code generation tasks in data science.  
This benchmark comprises 500 examples characterized by diverse data sources, complex task settings, and an executable environment. 
We develop DA-Agent, a robust LLM-Agent baseline, to tackle this challenging benchmark. However, experiments reveal that even the most advanced LLMs perform poorly on \ours, achieving only about a 30.5\% score. 
Future work will focus on 1) developing a more sophisticated data agent framework, 2) training more effective agents based on open-source LLMs.

%% file: sections/Limitations.tex
\section*{Limitations}

\ours introduces a challenging benchmark for agent code generation. The current version presents the following limitations:
While utilizing a substantial amount of data science data to fine-tune LLMs is meaningful, this approach has not been explored in this paper. Although this work proposes a general benchmark for data science, it warrants more thorough investigation. In future efforts, we plan to delve deeper into the performance of fine-tuning open-source LLMs on \ours.

%% file: appendix/examples.tex
\section{Task Examples}
\label{app:examples}

\begin{figure}[h]
    \centering
    \includegraphics[width=\textwidth]
    {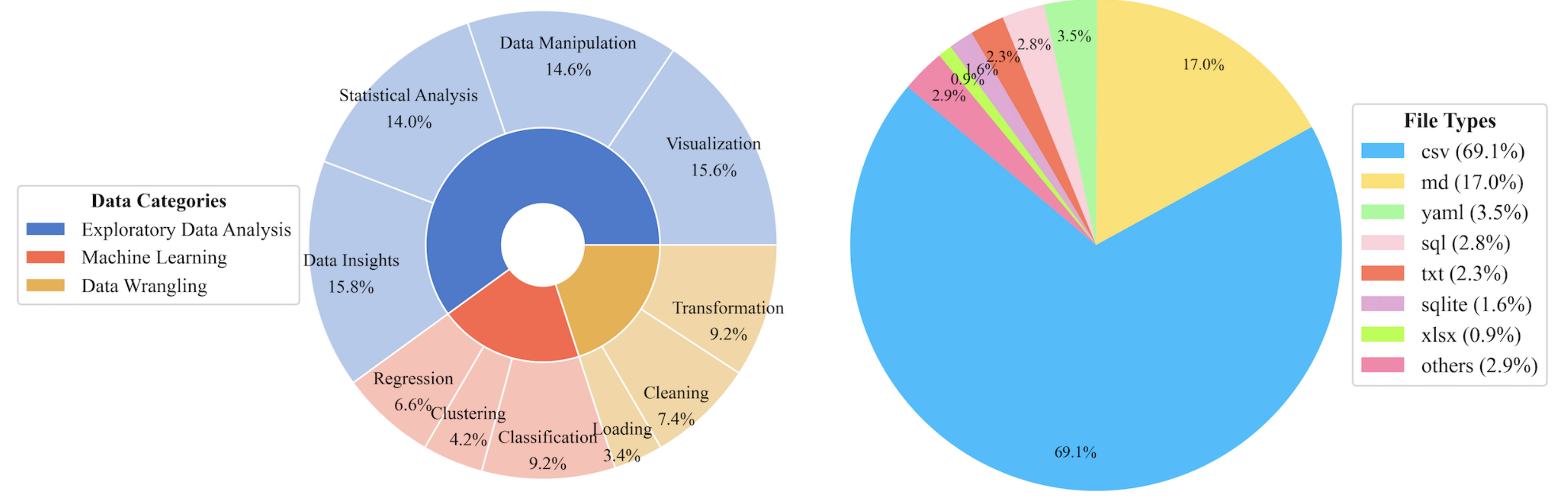}
    \caption{\ours task types proportion (Left) and file types proportion (Right)}
    \label{fig:statistic}
\end{figure}

In this section, we present diverse examples in \ours.
We have performed a more detailed classification of tasks into three categories: DW, ML, and EDA, as shown in Figure \ref{fig:statistic}. Data wrangling can be divided into data cleaning, data loading, and data transformation.
\setlist[itemize]{topsep=1pt, partopsep=0pt, parsep=0pt, itemsep=0pt}
\begin{itemize}
    \item \textit{Data Cleaning:} Focuses on enhancing the data's quality by eliminating errors, imputing missing values, normalizing data in databases or raw datasets, and resolving inconsistencies to maintain the accuracy and trustworthiness of the data.
    \item \textit{Data Loading:} Entails the consolidation of data from diverse sources into a unified storage system, loading data according to specified standards and requirements, enabling streamlined access and consistent analytical practices.
    \item \textit{Data Transformation:} Involves reformatting and restructuring data to better suit analytical models for targeted analysis.
\end{itemize}

The EDA category covers a wide range of tasks. In our benchmark, it is divided into \textit{Visualization, Statistical, Data Manipulation, and Data Insights}.
\setlist[itemize]{topsep=1pt, partopsep=0pt, parsep=0pt, itemsep=0pt}
\begin{itemize}
    \item \textit{Data Manipulation:} Data Manipulation is designed to perform intricate data operations using SQL and Pandas, efficiently processing large datasets. It primarily focuses on counting, summarizing data, and refining presentation formats.
    \item \textit{Data Insights:} This section focuses on tasks involving real data and issues encountered in actual scenarios. These problems do not provide clear solutions, necessitating thoughtful consideration and autonomous decision-making in coding for data analysis. The results are presented primarily in tables, or alternatively, in text format, to address and answer these questions.
    \item \textit{Visualization:} The format of visualization tasks is similar to the previous category, with the unique aspect being that the results are presented in the form of charts.
    \item \textit{Statistical Analysis:} Statistical analysis tasks typically require advanced knowledge of statistics and mathematics, utilizing mathematical indicators for statistical analysis.
\end{itemize}

ML tasks are categorized into three types: Classification, Regression, and Clustering. 
The reference solutions for these tasks require the use of corresponding ML algorithms to complete the assignments.
In \ours, we select three foundational task categories: classification, regression, and clustering, from two sources: regular dataset tasks and competition tasks.

Regular dataset tasks prioritize real-world applications, emphasizing comprehensive data preprocessing and feature engineering. 
Conversely, competition tasks present a heightened challenge, requiring advanced algorithmic approaches to meet specific performance metrics. 
In the instructions, the task category is not directly provided; the model needs to autonomously determine the task’s category, design methods, and select models accordingly. 
Additionally, only traditional machine learning algorithms, such as linear regression and random forests, are permitted, with no deep learning methods allowed. 
The final outputs consist of predictions on the test dataset provided.

\subsection{Data Wrangling Example}

\paragraph{Task Instruction}
\textit{According to the pre-defined database information, transform the data in the data folder and load it into the database.}

\paragraph{Verbose Instruction}\mbox{}\\
\textit{
1. Review the predefined schema to identify the tables and columns; note that there are five tables and record their column names.\\
2. Check which files are present in the ./data folder.\\
3. Examine the file information; since there are no minutes and hours information, data transformation is necessary.\\
4. Get trip\_minutes and trip\_hours by the trip\_duration.\\
5. The columns for rain and snow do not match with those in the raw data; type conversion is required.\\
6. Merge five months of JC data and compare with the schema to identify any missing columns.\\
7. Split the table JC by columns to match it with the database schema.\\
8. Write SQL insert statements to insert the data into the database.
}

\paragraph{Environment Settings}

\begin{verbatim}
|--- DATA_STANDARD.yml
|--- data
|    |--- JC-202401-citibike-tripdata.csv
|    |--- JC-202402-citibike-tripdata.csv
|    |--- JC-202403-citibike-tripdata.csv
|    |--- JC-202404-citibike-tripdata.csv
|    |--- JC-202405-citibike-tripdata.csv
|    `--- newark_airport_2024.csv
|--- database.db
`--- schema.yml
\end{verbatim}

\subsection{Machine Learning Task}
\paragraph{Task Instruction} 
\textit{This is a dataset for a Bank customer data for churn prediction competition, with the description available in README.md. You are now a contestant in this competition and need to design a method to predict the data in test.csv according to the competition requirements. Write the results into submission.csv according to the format of sample\_submission.csv.}

\paragraph{Verbose Instruction}
\textit{
1. Load the training, testing, and submission datasets from CSV files. 
}
\\
\textit{
2. Check the dimensions and basic statistics of each dataset, including the number of rows, columns, and presence of null values. 
}
\\
\textit{
3. Handle missing values in the datasets using appropriate methods such as imputation or removal. 
}
\\
\textit{
4. Scale numeric columns to ensure consistent ranges. 
}
\\
\textit{
5. Encode categorical text features using TF-IDF vectorization to transform them into numerical representations. 
}
\\
\textit{
6. Use a One-hot encoder to encode categorical features to convert them into a format suitable for machine learning models. 
}
\\
\textit{
7. Define feature columns for training the model, excluding non-predictive columns. 
}
\\
\textit{
8. Utilize CatBoostClassifier within a StratifiedKFold cross-validation framework to train and validate the model, ensuring robustness and performance assessment. 
}
\\
\textit{
9. Use the trained model to make predictions, and prepare the submission file by mapping predicted probabilities to the `Exited` column for submission. 
}
\paragraph{Environment Settings}
\begin{verbatim}
|--- README.md
|--- Churn_Modelling.csv
|--- train.csv
|--- test.csv
|--- sample_submission.csv
\end{verbatim}

\subsection{Exploratory Data Analysis (EDA)}

\subsubsection{Visualization}

\paragraph{Task Instruction}
\textit{Create a stacked horizontal bar chart, which illustrates the average days per order stage for the top 10 cities by sales. Save the chart as `result.png' with settings from `plot.yaml'.}

\paragraph{Verbose Instruction}\mbox{}\\
\textit{
1. Check Available Resources and Current Directory: View the resources provided and examine the contents of the current directory.\\
2. Database Content Review: Read what is contained in the database and identify the tables present.\\
3. Identify Top 10 Cities by Sales: To determine the top 10 cities by sales, join the `orders' and `customers' tables using the `customer\_id'. Record the names of these cities.\\
4.Create an SQL query to evaluate order processing times in the top 10 cities by joining the `orders' and `customers' tables using `customer\_id'. Calculate average durations for key milestones in the order process and include only orders from these cities, identified possibly via a subquery based on order volumes. Group and display results by `customer\_city', showing averages for each stage.\\
5. Read Plot Configuration: Load the `plot\_config.yml' file to review plotting requirements.\\
6. Create a Pie Chart of Average Order Processing Times: Prepare a summarized DataFrame, configure the pie chart with appropriate labels and colors, enhance its aesthetics with a title and legend, and then save and display the result. 
}

\paragraph{Environment Settings}

\begin{verbatim}
|--- DATASET_INFO.md
|--- E-commerce.db
`--- plot_config.yml
\end{verbatim}

\subsubsection{Data Manipulation }

\paragraph{Task Instruction}
\textit{
Utilize the Open Food Facts database. Identify the list of ingredients and their countries of origin, and record the results in the ingredient\_origins.csv file. 
}

\paragraph{Verbose Instruction}\mbox{}\\
\textit{
1. Read in the avocado data. Read the avocado data from a tab-delimited CSV file. Subset the DataFrame to include only a smaller number of relevant columns.Read in the relevant category tags for avocados from a text file.\\
2. Filter avocado data using relevant category tags - Drop rows with null values in the `categories\_tags' column. Convert the `categories\_tags' column from comma-separated strings to lists. Filter the DataFrame to keep only rows with relevant category tags.\\
3. Determine the top origin country for UK avocados. Filter the avocado DataFrame for rows where `countries' equals "United Kingdom". Count and order the unique values in the `origins\_tags' column. Identify the top country of origin for avocados in the UK. Lean up the country string to remove any leading characters or hyphens.\\
4. Create a user-defined function for ingredient analysis Create a function called `read\_and\_filter\_data()' that: Takes a filename and a list of relevant categories as arguments.Performs the same steps as above to read, subset, filter, and analyze the data. Returns the top country of origin for the ingredient.\\
5. Analyze other ingredients. Use the relevant categories data to determine the top origin countries for olive oil and sourdough by calling the `read\_and\_filter\_data()` function 
}

\paragraph{Environment Settings}

\begin{verbatim}
|--- avocado.csv
|--- ingredient_origins.csv
|--- olive_oil.csv
|--- project_instructions.md
|--- README.md
|--- relevant_avocado_categories.txt
|--- relevant_olive_oil_categories.txt
|--- relevant_sourdough_categories.txt
`--- sourdough.csv
\end{verbatim}

\subsubsection{Statistical Analysis}

\paragraph{Task Instruction}
\textit{
You have a Statistical thinking dataset, with details described in the README.md file. Calculate 10,000 bootstrap replicates of the variance in annual rainfall at the Sheffield Weather Station. Divide the data into 50 bins, compute the bin center and corresponding probability density function (PDF) for each bin. For convenience, convert the variance to units of square centimeters. Save the results to a file named result.csv, following the template provided in sample\_result.csv. (Set the random seed to 42)
}

\paragraph{Verbose Instruction}\mbox{}\\
\textit{
1. Bootstrap Helper Functions:\\
Define a function `bootstrap\_replicate\_1d(data, func)' to generate a bootstrap replicate of 1D data.\\
Define another function `draw\_bs\_reps(data, func, size=1)' to draw multiple bootstrap replicates.\\
2. Data Preparation:\\
Read the weather station CSV file considering it is space-delimited and does not have a header.\\
Assign appropriate column names.\\
Remove the first several rows if it contains non-numeric data.\\
Convert the year column to integers and rain column to floats, handling conversion errors gracefully and dropping any resulting NaN values.\\
Compute the total annual rainfall by grouping data by year and summing the rain values for each year.\\
Convert the resulting annual rainfall sums to a NumPy array.\\
3. Bootstrap Analysis:\\
Generate 10,000 bootstrap replicates of the variance of annual rainfall using the `draw\_bs\_reps' function.\\
Adjust the variance units if needed (e.g., put the variance in units of square centimeters).\\
4. Probability Density Function (PDF):\\
Create a histogram of the bootstrap replicates with 50 bins, normalized to represent a PDF.\\
Calculate the center points for each bin and the corresponding PDF values.\\
Store the bin centers and PDF values in a DataFrame.\\
5. Save Results.
}

\paragraph{Environment Settings}

\begin{verbatim}
|--- sheffield_weather_station.csv
|--- scandens_beak_depth_heredity.csv
|--- sample_result.csv
|--- mlb_nohitters.csv
|--- frog_tongue.csv
|--- fortis_beak_depth_heredity.csv
|--- finch_beaks_1975.csv
|--- finch_beaks_2012.csv
`--- README.md
\end{verbatim}

\subsubsection{Data Insights}

\paragraph{Task Instruction}
\textit{
What strategies could be implemented at electric vehicle charging stations to better accommodate the high volume of users and long-duration charging sessions observed at popular locations and peak times?
}

\paragraph{Verbose Instruction}\mbox{}\\
\textit{
1. Access Data Directory: Start by accessing the directory where data files are stored, including CSVs, documents, and images related to charging sessions.\\
2. Review Data Quality: Open and review files to check their formats (CSV, JSON, etc.) and data quality. Identify any missing values or inconsistencies.\\
3. Load Data: Import necessary files into an SQLite database.\\
4. Identify Popular Garages: Write a SQL query to count distinct users per garage, focusing on shared users to determine high-traffic garages.\\
5. Analyze Peak Times: Develop a SQL query to find peak charging times by analyzing sessions by day and hour.\\
6. Calculate Charging Durations: Create a SQL query to calculate average charging durations per user, focusing on those exceeding specific thresholds.\\
7. Analyze User Behavior: Identify patterns in long-duration charging by examining frequency and preferred times.\\
8. Segment User Data: If applicable, segment data by user type (commercial, personal, shared) to tailor improvements to different user needs.
}

\paragraph{Environment Settings}

\begin{verbatim}
|--- README.md
|--- analysis.md
|--- charging_sessions.csv
|--- historical_usage_2022.csv
|--- forecast_analysis_2023.md
|--- user_feedback_logs.md
|--- pricing_updates.json
`--- maintenance_records
     |--- january_2023.txt
     |--- february_2023.txt
     `--- march_2023.txt
\end{verbatim}

%% file: appendix/experiments_details.tex
\section{Experiments Details}
This section includes a detailed agent action space in Table \ref{tab:agent_actions}.




\input{tables/agent_actions}


%% file: tables/agent_actions.tex
\begin{table}[!h]
\centering
\begin{tabularx}{\textwidth}{lX}
\toprule
\textbf{Action Name} & \textbf{Action Description} \\
\midrule
File Viewing & The ability to access and review the contents of files, including but not limited to YAML (.yml) files, CSV (.csv) files, Markdown (.md) files, and text (.txt) files. \\
\midrule
File Operating & Performing operations on files such as creating, opening, editing, moving, and deleting. \\
\midrule
Data Processing & The manipulation and transformation of data to extract insights or prepare it for storage and further analysis. \\
\midrule
System Operations & Tasks related to the maintenance and management of computer systems, including monitoring performance, configuring settings, and troubleshooting issues. \\
\midrule
Package Management & The process of installing, upgrading, configuring, and managing software packages in an operating system. \\
\midrule
SQL Query & Executing read operations on databases using Structured Query Language (SQL) commands to retrieve specific data or summaries. \\
\midrule
SQL Update & Updating, modifying, or deleting data in a database using SQL commands, including insert, update, and delete operations. \\
\midrule
SQL Debug & The process of locating and resolving issues or bugs within SQL scripts or databases to ensure accurate data manipulation and retrieval. \\
\midrule
Python & Writing, executing scripts and programs in the Python programming language for a wide range of applications. \\
\midrule
Python Debug & Identifying and fixing errors or bugs within Python code to ensure correct execution and output of Python programs. \\
\midrule
Other & Activities or tasks that do not fall into the predefined categories, encompassing a miscellaneous range of actions. \\
\bottomrule
\end{tabularx}
\caption{Summary of DA-Agent actions with descriptions.}
    \vspace{-5pt}
    \label{tab:agent_actions}
\end{table}

%% file: appendix/evaluation_details.tex
\section{Evaluation details}
\label{app:evaluation_details}

\subsection{Plot based Evaluation}
Currently, existing methods for evaluating data visualizations include analyzing code correctness and using VLMs to score visualizations. The former relies on understanding the logic and structure of the code, which may lead to lower accuracy, while the latter has shown poor performance and high costs. 
We propose a more reasonable and reliable evaluation method for visualizations. This method involves embedding scripts into code and executing them to directly obtain data, types, and other information corresponding to the chart.

Specifically, given the following inputs: chart data \( d \), image metadata \( J \), the code used for generation, and the corresponding script. By embedding and executing the script within the generating code, we extract the predicted chart’s metadata \( \hat{J} \) and data \( \hat{d} \). The score is then calculated as follows:

\[
\text{Score} = \begin{cases}
1 & \text{if } d = \hat{d} \text{ and } J = \hat{J} \\
0 & \text{otherwise}
\end{cases}
\]

Here, if the task does not explicitly restrict the order or scale of the data, \( d' \) and \( \hat{d}' \) are the sorted or scaled versions of \( d \) and \( \hat{d} \). Additionally, \( j \subset J \) and \( \hat{j} \subset \hat{J} \), which are specific pieces of image information explicitly specified in the instructions, are compared from \( J \) and \( \hat{J} \), respectively. 

\subsubsection{Image Information Description}
To provide a comprehensive understanding, we will use an example to introduce all the relevant information involved in data visualization tasks in \ours.

\begin{figure}[htbp]
    \centering
    \includegraphics[width=0.9\textwidth]{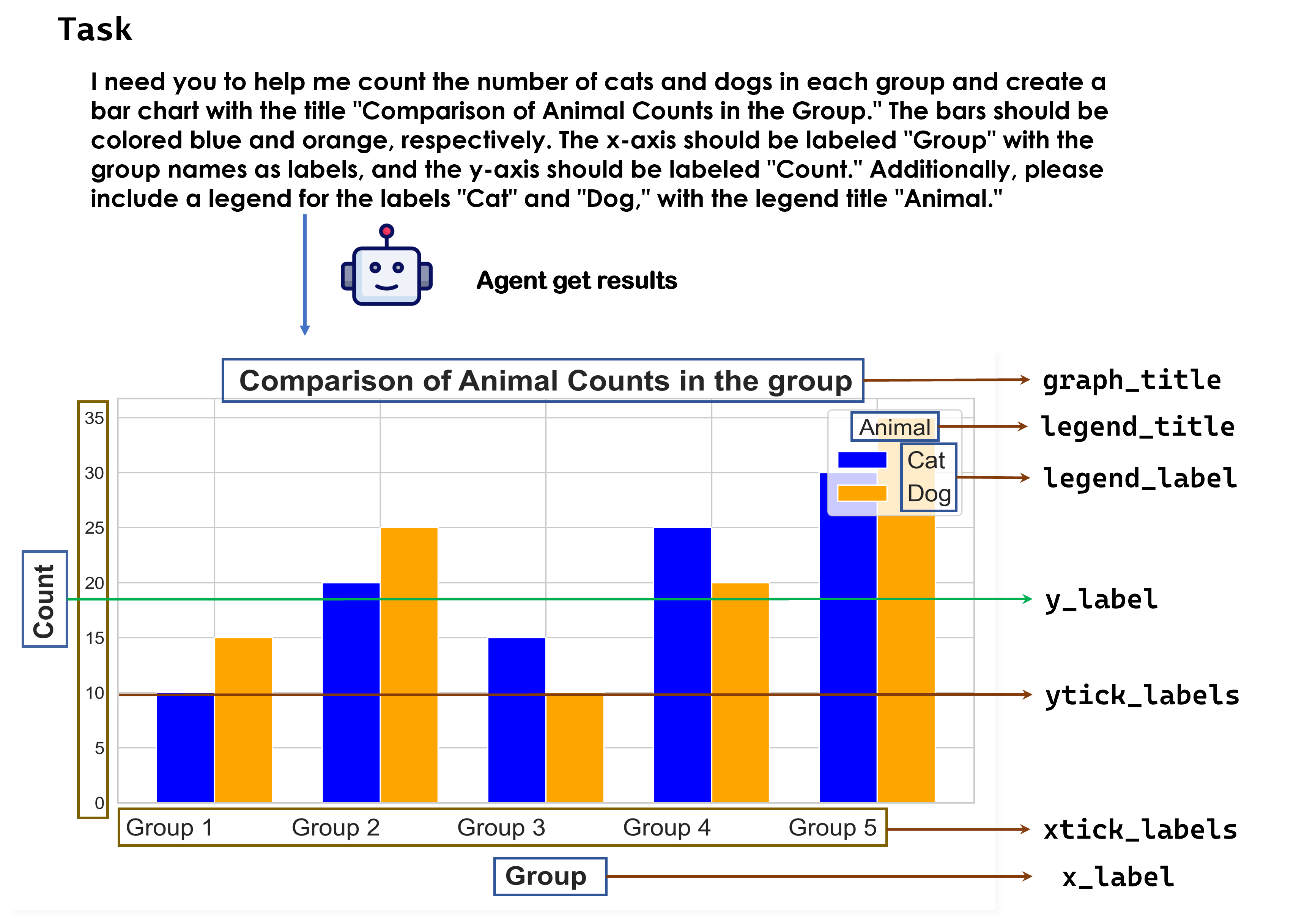}
    \caption{Example in data visualization task}
    \label{fig:visualization_example}
\end{figure}
\vspace{-10pt}
\paragraph{figsize} "figsize" specifies the shape or dimensions of the image. It determines the width and height of the plot in inches. In our example, it is set to (10, 6) (not explicitly shown in the image, but set in the code).
\vspace{-5pt}
\paragraph{color} "color" represents the color scheme used within the graph or visualization. It defines the hues applied to different data elements for differentiation or emphasis, like blue and orange in the example above.
\vspace{-5pt}
\paragraph{graph\_title} "graph\_title" provides the overarching title or caption for the graph. It succinctly summarizes the purpose or main findings depicted in the visual representation, for example, ``Comparison of Animal Counts in the group''.
\vspace{-5pt}
\paragraph{legend\_title} "legend\_title" describes the title of the legend, which typically explains the meaning of different colors or symbols used in the graph. It aids in understanding the data categories or groups represented, such as ``Animal''.
\vspace{-5pt}
\paragraph{labels} "labels" refers to the labels associated with the visualized data points or categories. These labels provide context for interpreting the data. In the example above, it is ``Cat'' and ``Dog''.
\vspace{-5pt}
\paragraph{x\_label} "x\_label" indicates the title for the x-axis, representing the horizontal dimension of the graph. It clarifies what the x-axis measures or represents, e.g., ``Group''.
\vspace{-5pt}
\paragraph{y\_label} "y\_label" represents the title for the y-axis, which denotes the vertical dimension of the graph. It clarifies what the y-axis measures or represents, e.g., ``Count''.
\vspace{-5pt}
\paragraph{xtick\_labels} "xtick\_labels" refers to the specific labels assigned to individual ticks or points along the x-axis. These labels provide scale and context to the data points plotted on the x-axis, e.g., ``Group 1'', ``Group 2'', ``Group 3'', ``Group 4'', ``Group 5''.
\vspace{-5pt}
\paragraph{ytick\_labels} "ytick\_labels" represents the specific labels assigned to ticks or points along the y-axis. Similar to xtick\_labels, these labels provide scale and context to the data points plotted on the y-axis, e.g., 0, 5, 10, 15, 20, 25, 30, 35.

 In this example, the task explicitly specifies the bar chart's colors, graph\_title, x\_label, xtick\_labels, y\_label, legend\_title, and labels. Therefore, during evaluation, we will filter out and compare these specific pieces of information.

\subsection{ML Task Evaluation}
To achieve consistent evaluation scores across machine learning tasks of varying difficulty levels and dataset sizes, we normalize evaluation metrics to a scale of 0-1 by setting basic and best bounds. 
The best bound is determined by the top solutions in the dataset and the highest-ranked metric in the competition. 
The basic bound is established using baseline methods: for classification tasks, the metric of a Proportional Classifier; for regression tasks, the metric of mean prediction; for clustering tasks, the metric of random clustering. 
In competition scenarios, the 80th percentile metric is used to determine the basic bound.

Given the best bound $S_{\text{best}}$, the basic bound $S_{\text{baseline}}$, and the predicted score $\hat{s}$ under the corresponding metric calculation, the scaled score is calculated as follows:

$$
Score = \min\left(\max\left(\frac{\hat{s} - S_{\text{baseline}}}{S_{\text{best}} - S_{\text{baseline}}}, 0\right
), 1\right) 
$$

This formula ensures that the score is normalized between 0 and 1, adjusting for whether higher or lower values indicate better performance.

Next, we will introduce the machine learning metrics used in \ours for classification, clustering, and regression tasks.

\subsubsection{Classification Tasks}
A detailed description of the metrics used in classification tasks, including their definitions and formulas, is provided below:

\paragraph{Accuracy}
Accuracy is defined as the proportion of correctly predicted instances out of the total instances:
\[ \text{Accuracy} = \frac{\text{Number of correctly predicted instances}}{\text{Total number of instances}} \]

\paragraph{F1 Score}
The F1 Score is a key evaluation metric for classification tasks, especially with imbalanced datasets. It is the harmonic mean of precision and recall, balancing the two. The F1 Score is calculated as follows:

\[ 
F1 = 2 \times \frac{\text{precision} \times \text{recall}}{\text{precision} + \text{recall}}
\]

Precision measures the accuracy of positive predictions, while recall measures the ability to identify all positive instances. 

\paragraph{Logarithmic Loss}
Logarithmic Loss (LogLoss) is a metric used to evaluate the performance of a classification model by measuring the accuracy of its predicted probabilities. It penalizes incorrect predictions, particularly those that are confidently incorrect. To avoid extremes, predicted probabilities are clipped using the formula \( \max(\min(p, 1e-15), 1 - 1e-15) \).

LogLoss can be computed in two primary ways:

One method involves averaging each class separately and then aggregating these averages. 
\[ 
\text{LogLoss} = -\frac{1}{M} \sum_{j=1}^M \frac{1}{N_j} \sum_{i=1}^{N_j} y_{ij} \log(p_{ij}) 
\]
where \( M \) is the number of classes, \( N_j \) is the number of samples in class \( j \), \( y_{ij} \) is a binary indicator (0 or 1) if class label \( j \) is the correct classification for sample \( i \), and \( p_{ij} \) is the predicted probability of sample \( i \) being in class \( j \).

Additionally, the other method involves averaging directly over all predictions without distinction by class. 
\[ 
\text{LogLoss} = -\frac{1}{N} \sum_{i=1}^N \sum_{j=1}^M y_{ij} \log(p_{ij}) 
\]
where \( N \) is the total number of samples, \( M \) is the number of classes, \( y_{ij} \) is a binary indicator (0 or 1) if class label \( j \) is the correct classification for sample \( i \), and \( p_{ij} \) is the predicted probability of sample \( i \) being in class \( j \).

\paragraph{ROC AUC Score} ROC AUC Score (Receiver Operating Characteristic Area Under the Curve) is a performance metric for binary classification tasks that evaluates the ability of a model to distinguish between classes. It measures the area under the ROC curve, which plots the true positive rate (sensitivity) against the false positive rate (1 - specificity) at various threshold settings.

\[ 
\text{ROC AUC Score} = \frac{1}{N_{\text{pos}} \cdot N_{\text{neg}}} \sum_{i \in \text{pos}} \sum_{j \in \text{neg}} I(\hat{y}_i > \hat{y}_j) 
\]

where \( N_{\text{pos}} \) and \( N_{\text{neg}} \) are the numbers of positive and negative samples respectively, \( \hat{y}_i \) and \( \hat{y}_j \) are the predicted scores for samples \( i \) and \( j \), and \( I(\cdot) \) is the indicator function that returns 1 if the condition is true and 0 otherwise.

\paragraph{Quadratic Weighted Kappa} Quadratic Weighted Kappa (QWK) is a metric used to assess the agreement between two ratings or annotations. It measures the extent of agreement beyond chance, considering both the magnitude of disagreement and its quadratic impact.

The formula for QWK is:
\[ 
\kappa = 1 - \frac{\sum_{i,j} w_{ij} O_{ij}}{\sum_{i,j} w_{ij} E_{ij}} 
\]
where \( O_{ij} \) is the observed agreement, \( E_{ij} \) is the expected agreement, \( w_{ij} = (i - j)^2 / (N-1)^2 \) represents the squared difference between ratings \( i \) and \( j \), and \( N \) is the total number of categories.

\subsubsection{Clustering Tasks}
The evaluation metric used for clustering tasks is introduced below.

\paragraph{Silhouette Score}
Silhouette score is a metric used to evaluate the quality of clustering in unsupervised learning. It measures how similar each sample is to its cluster compared to other clusters. The score ranges from -1 to 1, where a higher value indicates that samples are well-clustered, with tight clusters and distinct separation between them.

The formula for the Silhouette score for a single sample \( i \) is:

\[ 
s(i) = \frac{b(i) - a(i)}{\max(a(i), b(i))}
\]

where \( a(i) \) is the average distance from sample \( i \) to other samples within the same cluster (intra-cluster distance), and \( b(i) \) is the average distance from sample \( i \) to samples in the nearest neighboring cluster (inter-cluster distance).

The overall Silhouette score is the mean of \( s(i) \) for all samples in the dataset.

\subsubsection{Regression Tasks}
The evaluation metrics used for regression tasks are introduced below.

\paragraph{The Coefficient of Determination}The coefficient of determination (\( R^2 \)) is used to evaluate the proportion of variance in the dependent variable that is predictable from the independent variables. It indicates how well the regression predictions approximate the real data points.

The formula for \( R^2 \) is:

\[ 
R^2 = 1 - \frac{\sum_{i=1}^N (y_i - \hat{y}_i)^2}{\sum_{i=1}^N (y_i - \bar{y})^2}
\]

where \( N \) is the number of observations, \( y_i \) is the actual value, \( \hat{y}_i \) is the predicted value, and \( \bar{y} \) is the mean of the actual values. An \( R^2 \) value closer to 1 indicates a better fit of the model to the data.

\paragraph{Root Mean Squared Error}Root Mean Squared Error (RMSE) is used to evaluate the accuracy of regression models by measuring the square root of the average squared differences between predicted and actual values. It emphasizes larger errors due to squaring the differences.

The formula for RMSE is:
\[ 
\text{RMSE} = \sqrt{ \frac{1}{N} \sum_{i=1}^N (\hat{y}_i - y_i)^2 } 
\]
where \( N \) is the number of observations, \( \hat{y}_i \) is the predicted value, and \( y_i \) is the actual value. Lower RMSE values indicate better model performance, with a greater penalty for larger errors.

\paragraph{Root Mean Squared Logarithmic Error}Root Mean Squared Logarithmic Error (RMSLE) is used to evaluate the performance of regression models by measuring the logarithmic differences between predicted and actual values. It is particularly useful when the target variable spans several orders of magnitude.

The formula for RMSLE is:
\[ 
\text{RMSLE} = \sqrt{ \frac{1}{N} \sum_{i=1}^N \left( \log( \hat{y}_i + 1) - \log(y_i + 1) \right)^2 }
\]
where \( N \) is the number of observations, \( \hat{y}_i \) is the predicted value, and \( y_i \) is the actual value.

\paragraph{Mean Absolute Error}Mean Absolute Error (MAE) is used to evaluate the accuracy of regression models by measuring the average magnitude of errors in predictions, irrespective of their direction. It provides a straightforward measure of the average prediction error.

The formula for MAE is:
\[ 
\text{MAE} = \frac{1}{N} \sum_{i=1}^N | \hat{y}_i - y_i | 
\]
where \( N \) is the number of observations, \( \hat{y}_i \) is the predicted value, and \( y_i \) is the actual value. Lower MAE values indicate better model performance.

\paragraph{Median Absolute Error}Median Absolute Error (MedAE) is used to evaluate the accuracy of regression models by measuring the median of the absolute differences between predicted and actual values. It provides a robust metric that is less sensitive to outliers.

The formula for MedAE is:

\[ 
\text{MedAE} = \text{median}(|\hat{y}_i - y_i|)
\]

where \( \hat{y}_i \) is the predicted value and \( y_i \) is the actual value. Lower MedAE values indicate better model performance, with the metric focusing on the central tendency of errors.

\paragraph{Symmetric Mean Absolute Percentage Error}Symmetric Mean Absolute Percentage Error (SMAPE) is used to evaluate the accuracy of regression models by measuring the percentage error between predicted and actual values, treating over- and under-predictions equally.

The formula for SMAPE is:

\[ 
\text{SMAPE} = \frac{100\%}{N} \sum_{i=1}^N \frac{|\hat{y}_i - y_i|}{(|\hat{y}_i| + |y_i|) / 2} 
\]

where \( N \) is the number of observations, \( \hat{y}_i \) is the predicted value, and \( y_i \) is the actual value. Lower SMAPE values indicate better model performance, with the metric providing a balanced view of relative prediction errors.

%% file: appendix/prompts.tex
\section{Prompts of DA-Agent}
Following the ReAct \citep{yao2022react} prompting framework, which guides models in breaking down complex problems into manageable steps through an iterative chain of reasoning (Reasoning), action (Action), and observation (Observation) phases, where the current state is observed after each action before proceeding to the next step of reasoning, we build the DA-Agent. Below, we will introduce the system messages and actions used within this framework.

\subsection{System Prompt}

\begin{tcolorbox}
\begin{Verbatim}[breaklines=true]
# Context #
You are a data scientist proficient in analyzing data. You excel at using Bash commands and Python code to solve data-related problems. You are working in a Bash environment with all necessary Python libraries installed. If you need to install additional libraries, you can use the 'pip install' command. You are starting in the {work_dir} directory, which contains all the data needed for your tasks. You can only use the actions provided in the ACTION SPACE to solve the task. The maximum number of steps you can take is {max_steps}.
# ACTION SPACE #
{action_prompts}
# NOTICE #
1. You need to fully understand the action space and its arguments before using it.
2. You should first understand the environment and conduct data analysis on the given data before handling the task.
3. You can't take some problems for granted. For example, you should check the existence of files before reading them.
4. If the function execution fails, you should analyze the error and try to solve it.
5. For challenging tasks like ML, you may need to verify the correctness of the method by checking the accuracy or other metrics, and try to optimize the method.
6. Before finishing the task, ensure all instructions are met and verify the existence and correctness of any generated files.
# RESPONSE FROMAT # 
For each task input, your response should contain:
1. One analysis of the task and the current environment, reasoning to determine the next action (prefix "Thought: ").
2. One action string in the ACTION SPACE (prefix "Action: ").
# EXAMPLE INTERACTION #
Observation: ...(the output of last actions, as provided by the environment and the code output, you don't need to generate it)

Thought: ...
Action: ...

# TASK #
{task}
\end{Verbatim}
\end{tcolorbox}

\clearpage
\subsection{Action Prompts}

\paragraph{Bash Action}\mbox{}
\begin{tcolorbox}
\begin{Verbatim}[breaklines=true]
## Bash Action
* Signature: Bash(code="shell_command")
* Description: This action string will execute a valid shell command in the `code` field. Only non-interactive commands are supported. Commands like "vim" and viewing images directly (e.g., using "display") are not allowed.
* Example: Bash(code="ls -l")
\end{Verbatim}
\end{tcolorbox}

\paragraph{Python Action}\mbox{}
\begin{tcolorbox}
\begin{Verbatim}[breaklines=true]
## Python Action
* Signature: Python(file_path="path/to/python_file"):
```python
executable_python_code
```
* Description: This action will create a python file in the field `file_path` with the content wrapped by paired ``` symbols. If the file already exists, it will be overwritten. After creating the file, the python file will be executed. 
* Example: Python(file_path="./hello_world.py"):
```python
print("Hello, world!")
```
\end{Verbatim}
\end{tcolorbox}

\paragraph{SQL Action}\mbox{}
\begin{tcolorbox}
\begin{Verbatim}[breaklines=true]
## SQL Action
* Signature: SQL(file_path="path/to/database_file", command="sql_command", output="path/to/output_file.csv" or "direct")
* Description: Executes an SQL command on the specified database file. If `output` is set to a file path, the results are saved to this CSV file; if set to 'direct', results are displayed directly.
* Constraints:
  - The database file must be accessible and in a format compatible with SQLite (e.g., .sqlite, .db).
  - SQL commands must be valid and safely formatted to prevent security issues such as SQL injection.
* Examples:
  - Example1: SQL(file_path="data.sqlite", command="SELECT name FROM sqlite_master WHERE type='table'", output="directly")
  - Example2: SQL(file_path="data.db", command="SELECT * FROM users", output="users_output.csv")
\end{Verbatim}
\end{tcolorbox}

\clearpage
\paragraph{Terminate Action}\mbox{}
\begin{tcolorbox}
\begin{Verbatim}[breaklines=true]
## Terminate Action
* Signature: Terminate(output="literal_answer_or_output_path")
* Description: This action denotes the completion of the entire task and returns the final answer or the output file/folder path. Make sure the output file is located in the initial workspace directory.
* Examples:
  - Example1: Terminate(output="New York")
  - Example2: Terminate(output="result.csv")
  - Example3: Terminate(output="FAIL")
\end{Verbatim}
\end{tcolorbox}

%% file: appendix/trajectory_examples.tex
\section{Action Trajectory Examples}
This section provides action trajectory examples based on DA-Agent framework powered by GPT-4.

\begin{tcolorbox}
\begin{Verbatim}[breaklines=true]
## Example 1

* Task: This is a dataset titled “The-GitHub-History-of-the-Scala-Language,” with detailed descriptions provided in the README.md file. I want to know the number of pull requests made by two users with the nicknames “soc” and “xeno-by” for each year between 2011 and 2016. Please fill in your statistical results in the provided template file pull_requests_by_year_and_author.csv. 

[Step 1]
* Observation: You are in the folder now.

* Thought: Before I can analyze the data, I need to understand the structure of the dataset and the contents of the README.md file to know how the data is organized. I will use the `cat` command to display the contents of the README.md file.
* Action: Bash(code="cat README.md")
* code: cat README.md

[Step 2]
* Observation: # The-GitHub-History-of-the-Scala-Language
Project Description
---
Open source projects contain entire development histories - who made changes, the changes themselves, 
and code reviews. In this project, will be challenged to read in, clean up, and visualize the real-world 
project repository of Scala that spans data from a version control system (Git) as well as a project hosting 
site (GitHub). With almost 30,000 commits and a history spanning over ten years, Scala is a mature language. 
Will find out who has had the most influence on its development and who are the experts.

The dataset includes the project history of Scala retrieved from Git and GitHub as a set of CSV files.
file = 'src/compiler/scala/reflect/reify/phases/Calculate.scala'

* Thought: The README.md file indicates that the dataset consists of CSV files containing the project history of Scala from Git and GitHub. However, it does not specify the exact filenames or their contents. To proceed, I need to list the files in the current directory to identify the CSV files that might contain the pull request data.
* Action: Bash(code="ls -l")
* code: ls -l
\end{Verbatim}
\end{tcolorbox}

\begin{tcolorbox}
\begin{Verbatim}[breaklines=true]
[Step 3]
* Observation: total 4972
-rw-r--r-- 1 570063388 500000513     754 Dec 31  1979 README.md
-rw-r--r-- 1 570063388 500000513 4836135 Dec 31  1979 pull_files.csv
-rw-r--r-- 1 570063388 500000513      66 Jun  9 04:13 

pull_requests_by_year_and_author.csv
-rw-r--r-- 1 570063388 500000513  242097 Dec 31  1979 pulls.csv
* Thought: There are two CSV files that could contain the pull request data: `pull_files.csv` and `pulls.csv`. The `pull_requests_by_year_and_author.csv` file is likely the template I need to fill in with the statistical results. To determine which CSV file contains the pull request data and the information about the authors and dates, I should take a look at the first few lines of each CSV file. I will start by examining the `pulls.csv` file using the `head` command.
* Action: Bash(code="head pulls.csv")
* code: head pulls.csv

[Step 4]
* Observation: pid,user,date

163314316,hrhino,2018-01-16T23:29:16Z

163061502,joroKr21,2018-01-15T23:44:52Z

163057333,mkeskells,2018-01-15T23:05:06Z

162985594,lrytz,2018-01-15T15:52:39Z

162838837,zuvizudar,2018-01-14T19:16:16Z

162765232,howtonotwin,2018-01-13T04:37:17Z

162643301,hrhino,2018-01-12T14:29:49Z

162256029,mkeskells,2018-01-10T23:20:03Z

162197834,SethTisue,2018-01-10T18:31:25Z

* Thought: The `pulls.csv` file contains the columns `pid`, `user`, and `date`, which are the pull request ID, the username of the person who made the pull request, and the date the pull request was made, respectively. Since the task requires the number of pull requests made by the users "soc" and "xeno-by" for each year between 2011 and 2016, this file seems to contain the necessary information. I will now write a Python script to read this CSV file, filter the data for the two users and the specified years, and count the number of pull requests per year for each user. Then, I will fill in the results in the `pull_requests_by_year_and_author.csv` template file.
\end{Verbatim}
\end{tcolorbox}

\begin{tcolorbox}
\begin{Verbatim}[breaklines=true]
* Action: Python(file_path="./analyze_pulls.py"):
```python
import pandas as pd

# Load the pull requests data
pulls = pd.read_csv('pulls.csv')
# Convert the date column to datetime
pulls['date'] = pd.to_datetime(pulls['date'])

# Filter the data for the years 2011-2016 and for the users 'soc' and 'xeno-by'
filtered_pulls = pulls[(pulls['date'].dt.year >= 2011) & (pulls['date'].dt.year <= 2016) & (pulls['user'].isin(['soc', 'xeno-by']))]

# Group by year and user, then count the number of pull requests
pulls_by_year_and_user = filtered_pulls.groupby([filtered_pulls['date'].dt.year, 'user']).size().reset_index(name='count')

# Load the template file
template = pd.read_csv('pull_requests_by_year_and_author.csv')

# Merge the results with the template
result = template.merge(pulls_by_year_and_user, how='left', left_on=['year', 'author'], right_on=['date', 'user']).fillna(0)

# Drop the extra columns and save the result
result = result[['year', 'author', 'count']]
result.to_csv('pull_requests_by_year_and_author.csv', index=False)
```
* code: import pandas as pd

# Load the pull requests data
pulls = pd.read_csv('pulls.csv')

# Convert the date column to datetime
pulls['date'] = pd.to_datetime(pulls['date'])

# Filter the data for the years 2011-2016 and for the users 'soc' and 'xeno-by'
filtered_pulls = pulls[(pulls['date'].dt.year >= 2011) & (pulls['date'].dt.year <= 2016) & (pulls['user'].isin(['soc', 'xeno-by']))]

# Group by year and user, then count the number of pull requests
pulls_by_year_and_user = filtered_pulls.groupby([filtered_pulls['date'].dt.year, 'user']).size().reset_index(name='count')

# Load the template file
template = pd.read_csv('pull_requests_by_year_and_author.csv')
\end{Verbatim}
\end{tcolorbox}

\begin{tcolorbox}
\begin{Verbatim}[breaklines=true]
# Merge the results with the template
result = template.merge(pulls_by_year_and_user, how='left', left_on=['year', 'author'], right_on=['date', 'user']).fillna(0)

# Drop the extra columns and save the result
result = result[['year', 'author', 'count']]
result.to_csv('pull_requests_by_year_and_author.csv', index=False)

[Step 5]
* Observation: Traceback (most recent call last):
  File "/workspace/./analyze_pulls.py", line 19, in <module>
    result = template.merge(pulls_by_year_and_user, how='left', left_on=['year', 'author'], right_on=['date', 'user']).fillna(0)
  File "/usr/local/lib/python3.9/site-packages/pandas/core/frame.py", line 10832, in merge
    return merge(
  File "/usr/local/lib/python3.9/site-packages/pandas/core/reshape/merge.py", line 170, in merge
    op = _MergeOperation(
  File "/usr/local/lib/python3.9/site-packages/pandas/core/reshape/merge.py", line 794, in __init__
    ) = self._get_merge_keys()
  File "/usr/local/lib/python3.9/site-packages/pandas/core/reshape/merge.py", line 1310, in _get_merge_keys
    left_keys.append(left._get_label_or_level_values(lk))
  File "/usr/local/lib/python3.9/site-packages/pandas/core/generic.py", line 1911, in _get_label_or_level_values
    raise KeyError(key)
KeyError: 'year'

* Thought: The error indicates that the 'year' column is not present in the template file, which means I cannot merge on this column. I need to check the contents of the `pull_requests_by_year_and_author.csv` template file to understand its structure and correct the merge operation accordingly.
* Action: Bash(code="cat pull_requests_by_year_and_author.csv")
* code: cat pull_requests_by_year_and_author.csv

[Step 6]
* Observation: date,soc,xeno-by

2011,,

2012,,

2013,,

2014,,

2015,,

2016,,
\end{Verbatim}
\end{tcolorbox}

\begin{tcolorbox}
\begin{Verbatim}[breaklines=true]
* Thought: The template file `pull_requests_by_year_and_author.csv` has columns for 'date', 'soc', and 'xeno-by', with the 'date' column representing the year. The error occurred because I tried to merge on a 'year' column that does not exist in the template. Instead, I should use the 'date' column for merging. I need to adjust the Python script to correctly merge the data based on the 'date' column and fill in the counts for 'soc' and 'xeno-by' in their respective columns.
* Action: Python(file_path="./analyze_pulls.py"):
```python
import pandas as pd

# Load the pull requests data
pulls = pd.read_csv('pulls.csv')

# Convert the date column to datetime
pulls['date'] = pd.to_datetime(pulls['date'])

# Filter the data for the years 2011-2016 and for the users 'soc' and 'xeno-by'
filtered_pulls = pulls[(pulls['date'].dt.year >= 2011) &
(pulls['date'].dt.year <= 2016) & (pulls['user'].isin(['soc', 'xeno-by']))]

# Group by year and user, then count the number of pull requests
pulls_by_year_and_user = filtered_pulls.groupby([filtered_pulls['date'].dt.year, 'user']).size().reset_index(name='count')

# Pivot the result to have years as rows and users as columns
pivot_table = pulls_by_year_and_user.pivot(index='date', columns='user', values='count').reset_index().fillna(0)

# Rename the columns to match the template file
pivot_table.columns = ['date', 'soc', 'xeno-by']

# Load the template file
template = pd.read_csv('pull_requests_by_year_and_author.csv')

# Merge the results with the template
result = template.merge(pivot_table, on='date', how='left').fillna(0)

# Drop the extra columns and save the result
result = result[['date', 'soc', 'xeno-by']]
result.to_csv('pull_requests_by_year_and_author.csv', index=False)
```
* code: import pandas as pd

# Load the pull requests data
pulls = pd.read_csv('pulls.csv')

# Convert the date column to datetime
pulls['date'] = pd.to_datetime(pulls['date'])
\end{Verbatim}
\end{tcolorbox}

\begin{tcolorbox}
\begin{Verbatim}[breaklines=true]
# Filter the data for the years 2011-2016 and for the users 'soc' and 'xeno-by'
filtered_pulls = pulls[(pulls['date'].dt.year >= 2011) & (pulls['date'].dt.year <= 2016) & (pulls['user'].isin(['soc', 'xeno-by']))]

# Group by year and user, then count the number of pull requests
pulls_by_year_and_user = filtered_pulls.groupby([filtered_pulls['date'].dt.year, 'user']).size().reset_index(name='count')

# Pivot the result to have years as rows and users as columns
pivot_table = pulls_by_year_and_user.pivot(index='date', columns='user', values='count').reset_index().fillna(0)
# Rename the columns to match the template file
pivot_table.columns = ['date', 'soc', 'xeno-by']

# Load the template file
template = pd.read_csv('pull_requests_by_year_and_author.csv')

# Merge the results with the template
result = template.merge(pivot_table, on='date', how='left').fillna(0)

# Drop the extra columns and save the result
result = result[['date', 'soc', 'xeno-by']]
result.to_csv('pull_requests_by_year_and_author.csv', index=False)

[Step 7]
* Observation: Traceback (most recent call last):
  File "/workspace/./analyze_pulls.py", line 28, in <module>
    result = result[['date', 'soc', 'xeno-by']]
  File "/usr/local/lib/python3.9/site-packages/pandas/core/frame.py", line 4108, in __getitem__
    indexer = self.columns._get_indexer_strict(key, "columns")[1]
  File "/usr/local/lib/python3.9/site-packages/pandas/core/indexes/base.py", line 6200, in _get_indexer_strict
    self._raise_if_missing(keyarr, indexer, axis_name)
  File "/usr/local/lib/python3.9/site-packages/pandas/core/indexes/base.py", line 6252, in _raise_if_missing
    raise KeyError(f"{not_found} not in index")
KeyError: "['soc', 'xeno-by'] not in index"

* Thought: The error indicates that the columns 'soc' and 'xeno-by' are not found in the merged result, which suggests that the pivot operation did not produce these columns. This could be because the pivot operation did not find any data for these users in the specified years, or because the pivot operation was not performed correctly. I need to ensure that the pivot operation creates the columns for 'soc' and 'xeno-by' even if there are no pull requests for them in the data. I will modify the script to handle cases where there are no pull requests for these users by explicitly adding these columns if they are missing after the pivot operation.
\end{Verbatim}
\end{tcolorbox}

\begin{tcolorbox}
\begin{Verbatim}[breaklines=true]
* action: Python(file_path="./analyze_pulls.py"):
```python
import pandas as pd
# Load the pull requests data
pulls = pd.read_csv('pulls.csv')

# Convert the date column to datetime
pulls['date'] = pd.to_datetime(pulls['date'])

# Filter the data for the years 2011-2016 and for the users 'soc' and 'xeno-by'
filtered_pulls = pulls[(pulls['date'].dt.year >= 2011) & (pulls['date'].dt.year <= 2016) & (pulls['user'].isin(['soc', 'xeno-by']))]
# Group by year and user, then count the number of pull requests
pulls_by_year_and_user = filtered_pulls.groupby([filtered_pulls['date'].dt.year, 'user']).size().reset_index(name='count')

# Pivot the result to have years as rows and users as columns, filling missing values with 0
pivot_table = pulls_by_year_and_user.pivot(index='date', columns='user', values='count').fillna(0).reset_index()

# Ensure that 'soc' and 'xeno-by' columns exist, adding them if they don't
if 'soc' not in pivot_table:
    pivot_table['soc'] = 0
if 'xeno-by' not in pivot_table:
    pivot_table['xeno-by'] = 0
# Rename the columns to match the template file
pivot_table.rename(columns={'date': 'date'}, inplace=True)

# Load the template file
template = pd.read_csv('pull_requests_by_year_and_author.csv')

# Merge the results with the template
result = template.merge(pivot_table, on='date', how='left').fillna(0)

# Save the result
result.to_csv('pull_requests_by_year_and_author.csv', index=False)
```
* code: import pandas as pd

# Load the pull requests data
pulls = pd.read_csv('pulls.csv')

# Convert the date column to datetime
pulls['date'] = pd.to_datetime(pulls['date'])
\end{Verbatim}
\end{tcolorbox}

\begin{tcolorbox}
\begin{Verbatim}[breaklines=true]
# Filter the data for the years 2011-2016 and for the users 'soc' and 'xeno-by'
filtered_pulls = pulls[(pulls['date'].dt.year >= 2011) & (pulls['date'].dt.year <= 2016) & (pulls['user'].isin(['soc', 'xeno-by']))]

# Group by year and user, then count the number of pull requests
pulls_by_year_and_user = filtered_pulls.groupby([filtered_pulls['date'].dt.year, 'user']).size().reset_index(name='count')

# Pivot the result to have years as rows and users as columns, filling missing values with 0
pivot_table = pulls_by_year_and_user.pivot(index='date', columns='user', values='count').fillna(0).reset_index()

# Ensure that 'soc' and 'xeno-by' columns exist, adding them if they don't
if 'soc' not in pivot_table:
    pivot_table['soc'] = 0
if 'xeno-by' not in pivot_table:
    pivot_table['xeno-by'] = 0

# Rename the columns to match the template file
pivot_table.rename(columns={'date': 'date'}, inplace=True)

# Load the template file
template = pd.read_csv('pull_requests_by_year_and_author.csv')

# Merge the results with the template
result = template.merge(pivot_table, on='date', how='left').fillna(0)

# Save the result
result.to_csv('pull_requests_by_year_and_author.csv', index=False)
[Step 8]
* observation: ./analyze_pulls.py executed successfully. No output.

* thought: The Python script has executed successfully without any errors, which means the `pull_requests_by_year_and_author.csv` file should now be filled with the correct data. I will now check the contents of the updated CSV file to ensure that the data has been filled in as expected.
* action: Bash(code="cat pull_requests_by_year_and_author.csv")
* code: cat pull_requests_by_year_and_author.csv

[Step 9]
* observation: date,soc_x,xeno-by_x,soc_y,xeno-by_y
2011,0.0,0.0,12.0,20.0
2012,0.0,0.0,44.0,271.0
2013,0.0,0.0,117.0,123.0
2014,0.0,0.0,20.0,60.0
2015,0.0,0.0,24.0,3.0
2016,0.0,0.0,21.0,0.0
\end{Verbatim}
\end{tcolorbox}

\begin{tcolorbox}
\begin{Verbatim}[breaklines=true]
* thought: The output shows that the merge operation has created additional columns with suffixes `_x` and `_y`. This is likely because the merge operation found columns with the same name in both dataframes and automatically added suffixes to differentiate them.
I need to adjust the script to ensure that the merge operation does not create these additional columns and that the data is correctly placed under the 'soc' and 'xeno-by' columns as specified in the template.
* action :  Python(file_path="./fix_csv.py"):
```python
import pandas as pd

# Load the CSV file that needs fixing
data = pd.read_csv('pull_requests_by_year_and_author.csv')

# We only need the 'date' column and the counts for 'soc' and 'xeno-by'
# The correct counts are in the columns with suffix '_y'
# Rename '_y' columns to remove the suffix and drop '_x' columns
data.rename(columns={'soc_y': 'soc', 'xeno-by_y': 'xeno-by'}, inplace=True)
data.drop(columns=['soc_x', 'xeno-by_x'], inplace=True)

# Save the corrected data back to the CSV file
data.to_csv('pull_requests_by_year_and_author.csv', index=False)
```
* code: import pandas as pd

# Load the CSV file that needs fixing
data = pd.read_csv('pull_requests_by_year_and_author.csv')

# We only need the 'date' column and the counts for 'soc' and 'xeno-by'
# The correct counts are in the columns with suffix '_y'
# Rename '_y' columns to remove the suffix and drop '_x' columns
data.rename(columns={'soc_y': 'soc', 'xeno-by_y': 'xeno-by'}, inplace=True)
data.drop(columns=['soc_x', 'xeno-by_x'], inplace=True)
# Save the corrected data back to the CSV file
data.to_csv('pull_requests_by_year_and_author.csv', index=False)

[Step 10]
* observation: ./fix_csv.py executed successfully. No output.

* thought: The script to fix the CSV file has been executed successfully. Now, I should check the contents of the `pull_requests_by_year_and_author.csv` file again to confirm that the data is now correctly formatted according to the template.
* action: Bash(code="cat pull_requests_by_year_and_author.csv")
* code: cat pull_requests_by_year_and_author.csv
\end{Verbatim}
\end{tcolorbox}

\begin{tcolorbox}
\begin{Verbatim}[breaklines=true]
[Step 11]
* observation: date,soc,xeno-by
2011,12.0,20.0
2012,44.0,271.0
2013,117.0,123.0
2014,20.0,60.0
2015,24.0,3.0
2016,21.0,0.0

* thought: The contents of the `pull_requests_by_year_and_author.csv` file now correctly show the number of pull requests made by the users "soc" and "xeno-by" for each year between 2011 and 2016. The task is complete, and I can now provide the path to the updated CSV file as the final output.
* action: Terminate(output="pull_requests_by_year_and_author.csv")
* code:
\end{Verbatim}
\end{tcolorbox}